\RequirePackage{fix-cm}
\documentclass{svjour3}                      

\usepackage{cite}
\usepackage{amsmath,amssymb,amsfonts}
\usepackage{algorithmic}
\usepackage{graphicx}
\usepackage{float}  
\usepackage{subfigure}

\usepackage{textcomp}
\usepackage[flushleft]{caption}
\usepackage{verbatim}
\usepackage{enumerate}
\usepackage{stfloats}
\usepackage{booktabs}
\usepackage{multirow}
\usepackage{hyperref}
\hypersetup{hidelinks,
	colorlinks=true,
	allcolors=black,
	pdfstartview=Fit,
	breaklinks=true}
	
\usepackage{ulem}	

\usepackage{soul, color, xcolor}
\sethlcolor{green}
\soulregister{\cite}7 
\soulregister{\citep}7 
\soulregister{\citet}7 
\soulregister{\ref}7  
\soulregister{\pageref}7 
\soulregister{\eqref} 7
\soulregister{\textcomp}7
\soulregister{\amsmath}7
\soulregister{\amssymb}7
\soulregister{\amsfonts}7
\soulregister{\graphicx}7
\soulregister{\algorihmic}7

\usepackage{bbding}

\smartqed  
\usepackage{graphicx}
\begin{document}

\title{Group channel pruning and spatial attention distilling for object detection
}


\author{Yun Chu \and Pu Li \and Yong Bai \and Zhuhua Hu \and Yongqing Chen \and Jiafeng Lu 
         }


\institute{Yun Chu, Yong Bai, Zhuhua Hu, Yongqing Chen, Jiafeng Lu are with  School of Information and Communication Engineering, Hainan University, Haikou, 570288, China \\   \and
             Pu Li is with the School of Software and Microelectronics, Peking University, Beijing, 100871, China.\\
 Correspondence should be addressed to Yong Bai, E-mail: bai@hainanu.edu.cn
}

\date{Received: date 2021.09 / Accepted: 2022.03}

\maketitle
\begin{abstract}
Due to the over-parameterization of neural networks, many model compression methods based on pruning and quantization have emerged. They are remarkable in reducing the size, parameter number, and computational complexity of the model. However, most of the models compressed by such methods need the support of special hardware and software, which increases the deployment cost. Moreover, these methods are mainly used in classification tasks, and rarely directly used in detection tasks. To address these issues, for the object detection network we introduce a three-stage model compression method: dynamic sparse training, group channel pruning, and spatial attention distilling. Firstly, to select out the unimportant channels in the network and maintain a good balance between sparsity and accuracy, we put forward a dynamic sparse training method, which introduces a variable sparse rate, and the sparse rate will change with the training process of the network. Secondly,  to reduce the effect of pruning on network accuracy, we propose a novel pruning method called group channel pruning.  In particular, we divide the network into multiple groups according to the scales of  the feature layer and the similarity of module structure in the network, and then we use different pruning thresholds to prune the channels in each group. Finally, to recover the accuracy of the pruned network, we use an improved knowledge distillation method for the pruned network. Especially, we extract spatial attention information from the feature maps of specific scales in each group as knowledge for distillation. In the experiments, we use YOLOv4 as the object detection network and PASCAL VOC as the training dataset. Our method reduces the parameters of the model by 64.7$\%$ and the calculation by 34.9$\%$. When the input image size is 416×416, compared with the original network model with 256MB size and 87.1 accuracies, our compressed model achieves 86.6 accuracies with 90MB size. To demonstrate the generality of our method, we replace the backbone to Darknet53 and Mobilenet and also achieve satisfactory compression results.

\end{abstract}

\section{Introduction}

In recent years, CNNs (Convolutional Neural Networks) have become the dominant methods for various computer vision tasks, such as image classification \cite{sullivan2018deep}, object detection \cite{liu2020deep}, and  segmentation\cite{sultana2020evolution}.
The classification networks include AlexNet \cite{krizhevsky2017imagenet}, ResNet \cite{he2016deep}, MobileNets \cite{howard2018inverted}, and the object detection networks include Faster-RCNN \cite{ren2016faster}, SSD \cite{liu2016ssd}, YOLOv3 $\sim $ v4 \cite{farhadi2018yolov3}, \cite{bochkovskiy2020yolov4}. The neural network models for those tasks have evolved from 8 layers to more than 100 layers.

Though the large networks have strong feature representation ability, they consume more resources. As an example, the depth of the YOLOv4 network reaches 162 layers, the size of the model is 256 MB, and the number of parameters is 64 million. When processing a picture of 416×416 size, it needs 29G FLOPs (Floating Point Operations Per Second), and the intermediate variables will occupy more memory. Taking into account the size of the model, the memory needed for inferencing and the amount of computation  are unbearable for resource-limited embedded devices.

To address the deployment problem of neural networks in mobile or embedded equipment, many model compression works  are based on pruning, quantification, knowledge distillation, and lightweight network design methods. In the pruning work, the pruning methods based on weight level were proposed \cite{han2016eie} and \cite{rathi2018stdp} to reduce the number of parameters of the model without affecting the accuracy of the network. However, the pruned models based on weight level requires special hardware accelerators to be deployed, such as \cite{abderrahmane2020design}. To save the cost of deployment, pruning methods based on filter level were proposed in \cite{luo2020autopruner}, \cite{fernandes2021pruning}, and these methods will not require special hardware support. In the quantification work, the binary network and ternary network were proposed in \cite{cheng2021intelligent} and \cite{deng2018gxnor}, respectively.
In  \cite{tung2018deep} and \cite{hu2021opq},  they combine the pruning with quantization and apply it to the classification network. In the work \cite{hu2021opq}, the information of pre-trained parameters is used to assign the compression ratio of each layer, and the method of the shared codebook is used to quantify and they achieve  a good compression effect. Though the low-bit quantification network can reduce the size of the model, they bring great accuracy loss and usually need a special software acceleration library to support deployment. The above pruning and quantization investigations are mainly used in classification networks, and not much has been examined for applying them to detection networks.

Pruning and quantization are  to compress the existing network structure and parameters. In contrast, knowledge distillation and lightweight network design optimize or directly design a new network structure, that is avoid the accuracy loss caused by pruning or quantization. Knowledge distillation is an  approach to improve the performance of the student network by using the teacher network. Hinton first applied the idea of distillation to the classification network \cite{hinton2015distilling}.Afterwards, knowledge distillation has been widely used in computer vision \cite{xu2019lightweightnet},\cite{zhang2021adversarial},\cite{song2021classifier}, natural language processing \cite{wang2021joint}, speech recognition \cite{li2021mutual}. Although knowledge distillation can improve the performance of the student network to a certain extent, its effect on reducing the parameters and size of the model is far from pruning. Moreover, how to represent the knowledge to be distilled between the teacher network and the student network is a problem. The structural difference between the teacher and  the student network has a great influence on the distillation effect. In \cite{shen2020knowledge} knowledge distillation is combined with representation learning to reduce the influence of structural difference on distillation. The work \cite{yang2021partially}  gives a reference to use the representation learning to solve the problem of partial view alignment.  \cite{ibrahem2021real} present a weakly supervised in object detection, this method require only image labels and counts of the objects of each class in the image, by this combination produce a clear localization of objects in the scene through a masking technique between class activation maps  and regression activation maps,  these work may solve the problem of knowledge representation in distillation to some extent. Lightweight network design is a direct design of small networks or modules, the article \cite{zhou2021rsanet} presents a lightweight network application in object detection tasks, including attention feature module to improve network accuracy, constant channel module to save memory access costs. Two different encoder-decoder lightweight architectures for semantic segmentation tasks were proposed in \cite{zhou2022contextual} and \cite{zhou2020aglnet}, respectively. The former work up-samples the convolution features of deep layer to the shallow deconvolution layers to enhance the contextual cues, and the later work uses channel split and shuffle modules in the encoder to reduce the number of parameters, and introduces an attention module in the decoder to improve the accuracy.  By directly designing lightweight modules, a good balance between the accuracy and the size of the model can be achieved, these lightweight networks can be well combined with tasks in autonomous driving, like\cite{kim2021edge}. Such these well-designed modules or networks \cite{sandler2019mobilenetv2}, \cite{litiny}, \cite{qin2019thundernet}, \cite{wang2018pelee}, \cite{liupath},  can normally run on laboratory host machines. However, if we want to deploy these networks successfully on edge devices, a lot of experiments and modifications are needed to verify their effectiveness.  Before the network is deployed to the edge device, the parameters and network need to be quantified and compiled. Usually, some innovative modules or network layers cannot be compiled and passed (due to the limitation of the instruction set and basic operators on the hardware device), which hinders the deployment of these lightweight networks to edge devices and reduces the versatility of these modules.

In the object detection task, model compression is mainly realized by knowledge distillation, and there exists little work to combine pruning with knowledge distillation. In short, the existing model compression works have the following limitations: 1. Most pruning and quantization models need special hardware circuits or software acceleration library support, which increases the cost to deploy these models to the edge device. 2. In the object detection networks, the model compression methods are mainly realized by knowledge distillation, and the compression effect is not satisfactory. Pruning is widely used in the classification network but  directly applied in  object detection.

To tackle the above problems, we propose a three-stage model compression method for object detection tasks: dynamic sparse training, grouping channel pruning, and spatial attention distillation. As shown in {Fig.1}, we briefly describe the implementation of the proposed three-stage model compression method.

\begin{figure*}[htbp]
\centering
\includegraphics[width=0.99\textwidth]{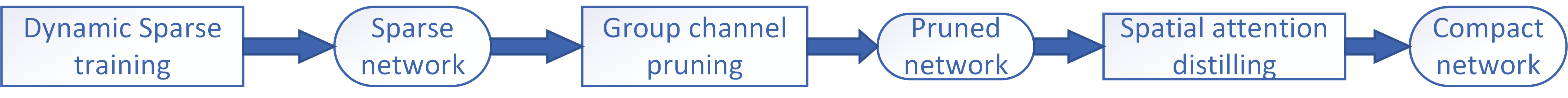}

\caption{Flow-chart of three-stage model compression: dynamic sparse training, group channel pruning, and spatial attention distilling.}
\label{Fig.1}
\end{figure*}

Firstly, we sparsely train the network. Sparse training is to make the distribution of $\gamma$  coefficient in the BN layer close to 0, and then the value of $\gamma$  coefficient is used as the importance scale factor of the channel to select out the insignificant channels in the network.  The traditional sparse training method uses a constant sparse rate in the training process, which is time-consuming and difficult to make a good balance between the sparsity and accuracy of the network. Therefore, we introduce a variable sparse rate to accelerate the sparse training of the network  and achieve a good balance between the sparsity and accuracy of the network, details are in Section 4.1  

Next, we prune the network. Most of the traditional pruning methods are used in classification networks, and all channels in the network are pruned with the same threshold. In contrast, we divide the detection network into multiple groups. In grouping, we mainly consider the scale of feature layers and the similarity of module structure in the network, the feature layers with the same scale and the layers have similar module structure are assigned to the same one group. After that, each group obtains the pruning threshold according to the current group's pruning ratio, then we prune the channels according to the pruning threshold in each group, thus achieving more accurate and efficient pruning of the detection network, details are in Section 4.2. 

At last, when pruning the channel of the detection network, we notice that with the increase of the detection category and pruning ratio, pruning will bring greater accuracy loss to the model. To recover the accuracy of the network after grouping pruning, we introduce knowledge distillation to the pruned network. Particularly, we extract spatial attention information from the feature maps of specific scales in each group as knowledge and distill the pruned network, details are in Section 4.3.

To the best of our knowledge, the work to combine pruning with spatial attention distillation and apply it to object detection tasks is currently rarely explored. The main contributions of this paper are summarized as follows:

\begin{itemize}
  \item [1)] 
To improve the efficiency of sparse training, we design a dynamic sparse training method which use variable sparsity rate to accelerate the process of sparse training, the network achieves a better trade-off between sparsity and accuracy.

  \item [2)]
For the object detection network, we propose a novel pruning method, called group channel pruning. We divide the detection network into multiple groups. During the group, we comprehensively consider the scale of feature layers and the similarity of module structure in the network. After that, each group obtains the pruning threshold  according to different pruning ratios, then we prune the channels in each group to achieve more accurate and efficient pruning of the detection network.

  \item [3)]
To recover the accuracy of the pruning network model, we introduce knowledge distillation to the network after grouping pruning.  In particular, we extract spatial attention information only from the feature maps of specific scales in each group as knowledge  and distill the pruned network. Furthermore, we demonstrate that our distillation method is not only suited for our pruning method but also can combine with other common pruning methods.  

  \item [4)]
We conduct extensive experiments on the PASCAL VOC data set with the YOLOV4 network to verify the effectiveness of the proposed method. To demonstrate the generality of our method, we also use Darknet53 and Mobilenet as the backbone to construct the detection network, the experimental results show that our method has other applicability.  In addition, we deploy the compression model on the edge device Jetson nano, which proves that our compression model can be deployed without special hardware support and can achieve an acceleration effect.

\end{itemize}

\section{Related work}
In this section, we briefly review the related works of pruning and knowledge distillation.

\subsection{Network Pruning}
The idea of pruning is to reduce the redundancy of structure and parameters in neural networks so that the network becomes more lightweight and efficient. The research of pruning focuses on two aspects. One is what kind of objects in the network can be pruned, and the other is how to measure the importance of the content being pruned. From the perspective of the object being pruned, the current pruning method can be divided into unstructured pruning and structured pruning. Unstructured pruning refers to that the topological structure of the network becomes irregular and unstructured after the network is pruned, and they often prune the connection weights between neurons. For example, in \cite{han2016eie} and \cite{rathi2018stdp}, the absolute value of the weight is taken as the metric of its pruning. The advantage of unstructured pruning is that the pruning rate can reach a high level without affecting the precision of the network. The disadvantage is that it needs the support of special hardware circuits, which increases the cost of deployment.

Structured pruning means that the network topology has not changed  after pruning. Usually pruning at the filter \cite{luo2017thinet}, channel \cite{liu2017learning}, layer \cite{jordao2020discriminative} levels. 
The work in \cite{luo2017thinet} prunes the unimportant filter in the current layer by calculating the statistical information of the subsequent layer.  Liu et al \cite{liu2017learning} propose a structured channel pruning method for classification networks, and the compressed model does not require special hardware and software support. The work in \cite{jordao2020discriminative} uses  the subspace projection method to measure the importance of the network layer, prunes the layer in the network, and verifies that the layer pruning is better than the filter pruning in resource utilization.  The advantage of  the structured pruning method is that the network after pruning does not need the support of special hardware circuits, and the disadvantage is that the pruning rate cannot reach very high.

Rethinking the Value of Network Pruning \cite{liu2018rethinking} discussed the significance of network pruning. Their work pointed out that the role played by pruning was similar to that of network architecture search (NAS). After that,  \cite{li2020eagleeye} bring the idea of NAS into pruning, proposes a batch normalization module to measure the importance of each structure in the network, and uses this NAS pruning method to the classification network. In this paper, we inherit the idea of sparse training in \cite{liu2017learning}, different from that, we introduce variable sparse rate to conduct dynamic sparse training on the network, which improves the trade-off between sparsity and accuracy of the network.

\subsection{Knowledge distillation}
The purpose of knowledge distillation is to transfer the knowledge learned from the teacher network to the student network to improve the performance of the student network. The research of knowledge distillation focuses on two aspects. One is which object in the network is selected as knowledge. The other is how to measure whether the student network learns the knowledge, which is reflected in how to design the loss function of distillation. Concerning what to be selected as knowledge, current distillation methods can be divided into three categories: 1. Using the final class information output of the teacher network as knowledge as in \cite{zhou2020two}, \cite{jung2020knowledge}, \cite{chen2018training}. 2. Using the middle feature layer of teachers' network as knowledge as in \cite{chen2017learning}, \cite{komodakis2017paying}. 3. Using the structural relationship between the layers of teacher network as knowledge as in \cite{wang2021joint}. In the classification network, \cite{komodakis2017paying} proposed to extract the attention from the feature layer and express the attention information in a heat map. Then, the loss function is constructed using the attention of the teacher network and the student network.

Search to Distill: \cite{liu2020search} introduced the knowledge distillation method into NAS, and obtained the following conclusions through experiments, the structure of student network determines the upper bound that the distillation effect can reach, and the distillation effect is better when the network structure of students and teachers is similar. Inspired by the above research work, we combine pruning with knowledge distillation. In this paper, the idea of our distillation is inspired by \cite{komodakis2017paying}, and we improved the method to make it suitable for object detection. Especially, we extract spatial attention from the feature layers of each group and give each group spatial attention with different weights for distilling.

\section{Network Architecture}
This paper takes YOLOv4 as an example to illustrate our pruning and distillation methods.
Our pruning method can be applied to networks with BN modules. In this section, we briefly introduce the five core basic components of the network and the overall architecture of the network. 

\subsection{Basic Components}
As shown in {Fig.2}, the five modules are CBM, Res Unit, CSP X, CBL, and SPP. 

\begin{figure}[H]
\begin{center}
\includegraphics[width=0.65\textwidth]{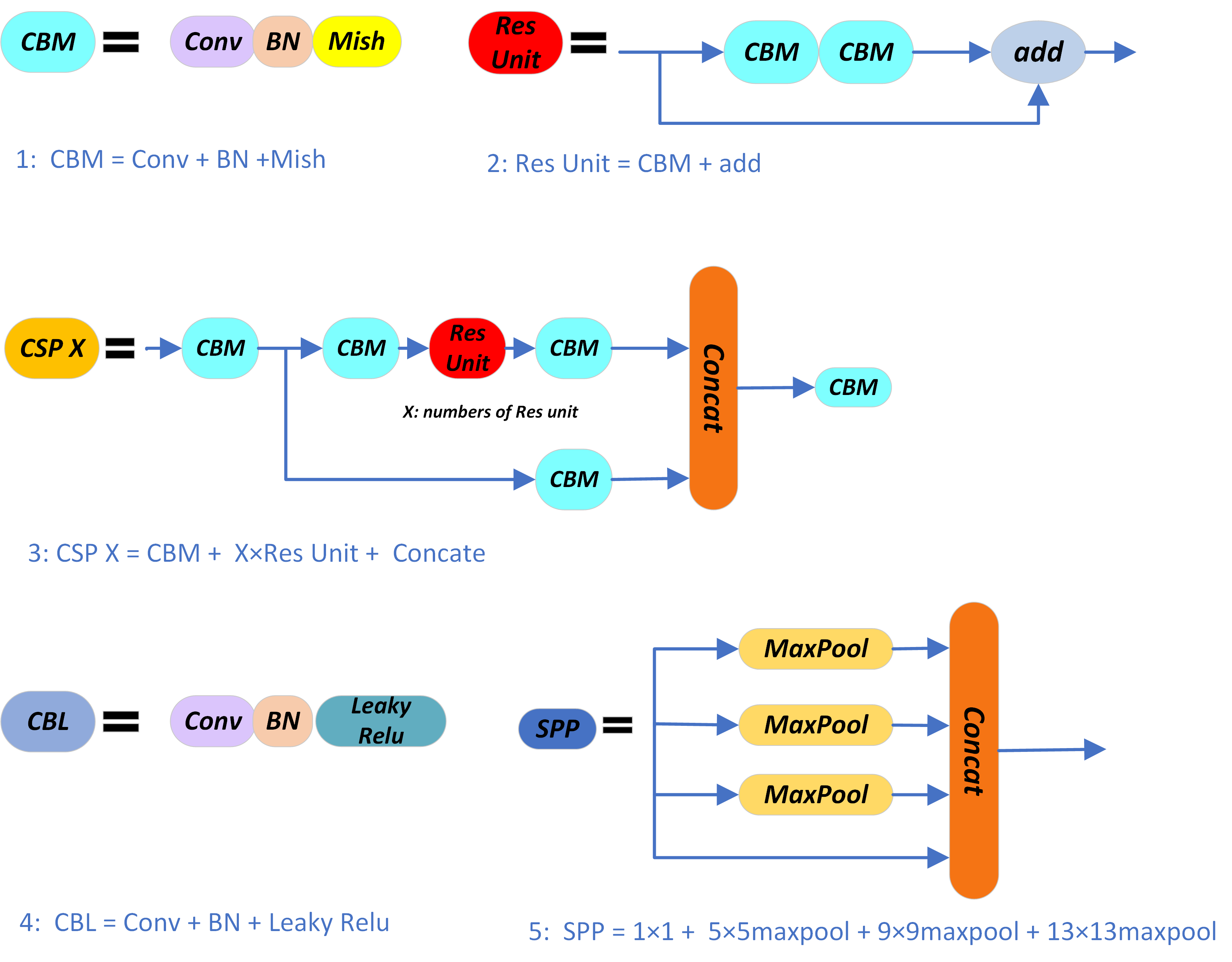}    
\end{center}
\caption{The five basic modules of YOLOv4 Network.}
\label{Fig.2}
\end{figure}

Among them, the CBM module is composed of Conv, Batch Normalization \cite{santurkar2018does} and Mish activation \cite{dasgupta2021performance} function. Res Unit is composed of CBM and add operations, where the res block is derived from \cite{he2016deep}. CSP X module is composed of CBM, X Res units, and concatenate operations. CBL module is composed of Conv, BN, and Leaky-Relu \cite{liu2019modified}. Spatial pyramid pooling (SPP) is proposed in \cite{huang2020dc}, herein SPP refers to feature fusion by pooling at four scales: 1×1, 5×5, 9×9, 13×13.

\subsection{Detection Network}
Through the above five basic components constitute the three parts of YOLOv4, i.e.,  backbone network, feature enhancement network, and detection head. These three parts have a total of 162 layers. Firstly, the backbone network is used to extract the feature of the object. Then, the feature enhancement network further fuses and enhances the features. At last, the detection head is responsible for classifying the input features and returning the location and size of the target. As shown in {Fig.3}m when the input size is 416 × 416, the down-sampling feature maps of 8, 16, 32 times are obtained through the backbone network, specifically, feature maps at scales 52×52, 26×26, 13×13, and then these feature maps are fed into the feature enhancement network. Finally, the detection head outputs the prediction at three scales.

\begin{figure}[H]
\begin{center}
\includegraphics[width=0.95\textwidth]{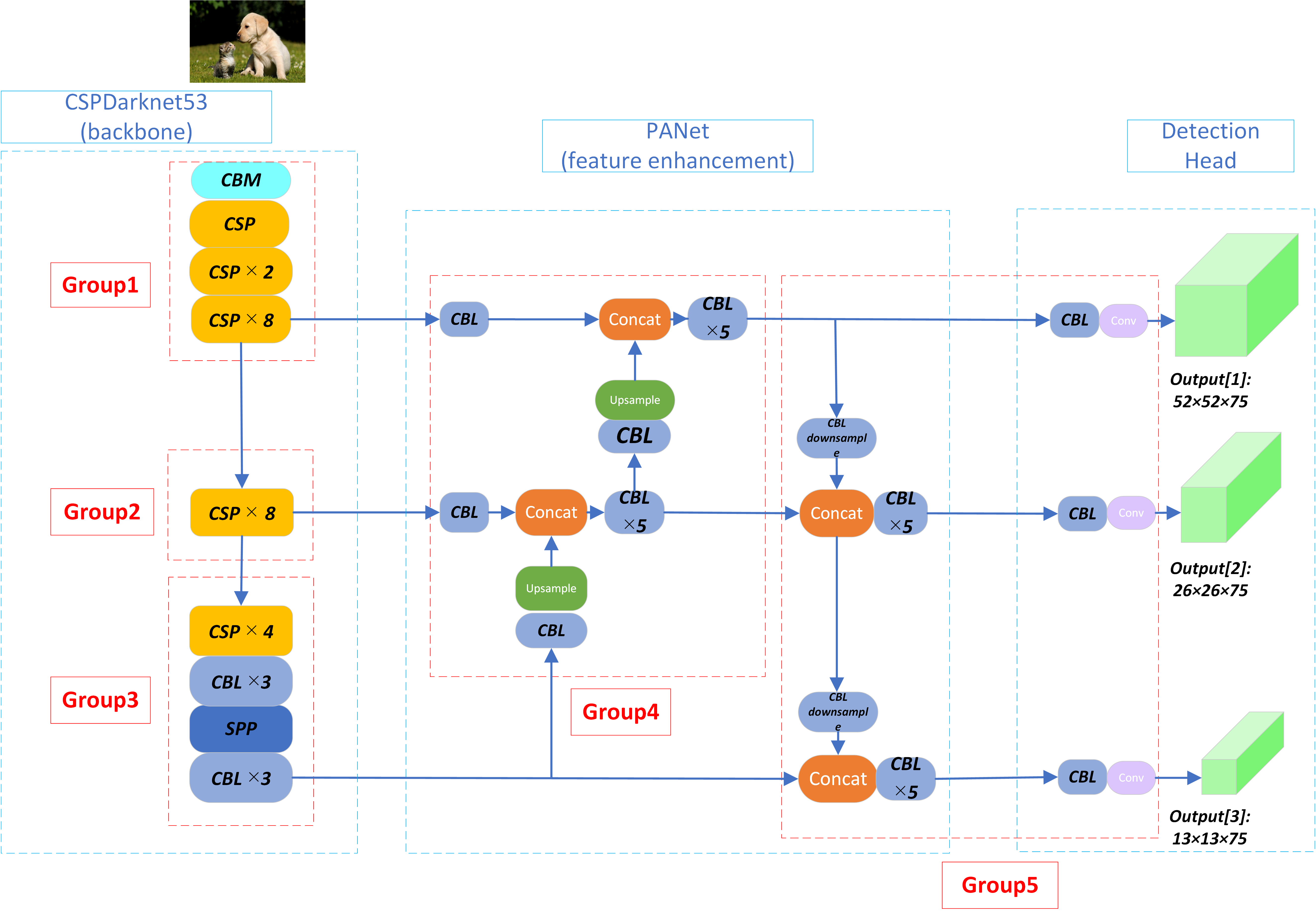}
\end{center}
\caption{Dividing YOLOv4 into five groups. In the graph, the blue dotted box contains three main parts of the network, the red dotted boxes represent the five groups. When grouping, we comprehensively consider the scale of feature layers and the similarity of module structure in the network. For example, when the input image size is 416×416,  group1: including the scales of feature layer from 416×416 to 52×52 and CSP module, group2: only include the feature layer at 26×26 scales along with the CSP module, group3:only include the feature layer at 13×13 scales along with the CSP and CBL module, same for the group4 and group5. Notice that, CSP module includes the CBM module as shown in {Fig.2}, the SPP,  Concat, Upsample, and Downsample these modules are just doing the calculation, they do not contain the trainable parameters. }
\label{Fig.3}
\end{figure}

\section{Proposed Approach}
In this section, we describe the proposed three-stage model compression method in detail. Firstly, we train the network sparsely with the dynamic sparse rate. Then, the object detection network is divided into five groups, and each group uses different thresholds for channel pruning. After that, using the pruned network as the student network for knowledge distillation, the details are described as follows.

\subsection{Dynamic Sparse Training}
The purpose of sparse training is to select out the insignificant channels in the network layer. Referring to the\cite{liu2017learning} , we use $\gamma$ as the important factor of the channel. The distribution of $\gamma$ coefficients of all BN layers in the original network is in different ranges. Sparse training is to sparse the $\gamma$ coefficient, making the distribution of the $\gamma$ coefficient close to zero. The smaller  $\gamma$ value  indicates the lower importance of the corresponding channel.
As shown in \eqref{1}, $\gamma$ is the scale parameter of the BN layer, $\beta$ is the offset parameter of the BN layer, the value of $\gamma$ and $\beta$ are obtained by training the network.  

\begin{equation}
\mathrm{y}_{i}=\gamma \hat{x}_{i}+\beta, \quad \hat{x}_{i}=\frac{x_{i}-\mu_{B}}{\sqrt{\sigma_{B}^{2}+\varepsilon}}, \label{1}
\end{equation}

where $\hat{x}_{i}$ denotes the normalized output of a channel, and ${y}_{i}$ denotes the output of $\hat{x}_{i}$ after $\gamma$ scaling and $\beta$ translation. ${x}_{i}$ denotes a specified channel on the feature layer. As shown in \eqref{2}, $\mu_{B}$ is the mean value of the specified channel under batch-size number, $\sigma_{B}^{2}$ is the variance of the specified channel under batch-size number. To prevent denominator being 0, $\varepsilon$ can be set a value at 1e-16. 

\begin{equation}
\mu_{B}=\frac{1}{m} \sum_{i=1}^{m} x_{i}, \quad \sigma_{B}^{2}=\frac{1}{m} \sum_{i=1}^{m}\left(x_{i}-\mu_{B}\right)^{2}, \label{2} 
\end{equation}

In the process of dynamic sparse training, the L1 norm of $\gamma$ is used as the regularization term, and the variable sparse rate $\mathrm{s}_{d}$ is introduced, which is added to the loss function for training, as shown in \eqref{3}.

\begin{equation}
\mathrm{L}=\sum_{(x, y)} l(f(x, W), y)+\mathrm{s}_{d} \sum_{\gamma \in \Gamma} g(\gamma), \label{3} 
\end{equation}

where $(x,y)$ represents the input of the network and the label of the data, $W$ represents the parameters that can be trained, and the first summation term represents the original loss during CNN network training. The second summation, $ g(\gamma)$ represents the regularization term introduced, we use $\mathrm{g}(\gamma)=|\gamma|$. $\mathrm{s}_{d}$ is a variable sparse rate. In the training process, the network will dynamically adjust its sparse rate according to the number of epochs of current training. When training to half the number of epochs, 70 $\%$ of the channel maintains the original sparse rate, 30 $\%$ of the channel sparse rate decay to 1 $\%$ of the original sparse rate, so that the final training network achieves a good balance between sparsity and accuracy.

\subsection{Group channel pruning}
In this section, we focus on the proposed group channel pruning method. It can be divided into three steps. Firstly, we divide the structure of the object detection network into five groups. Secondly, we obtain five different pruning thresholds according to the pruning proportion of each group and then generate the pruning mask matrix for most of the convolution layer in each group, which is used to prune the channels in the convolution layer. At last, we generate the public pruning mask matrix for the convolution layers associated with the shortcut layers.

\subsubsection{Network group}
Our grouping channel pruning is to divide the detection network layers into multi groups. In grouping, we comprehensively consider the scale of feature layers and the similarity of module structure in the network which means that the feature layers with the same scale and the layers that have similar module structure are assigned to the same group. In the experiment, we observe that when the unified pruning threshold is used for all structures, two detrimental situations will occur. One is that in structures with high redundancy, the real pruning threshold is higher than the unified pruning threshold, and the redundant channels in such structures will not be pruned. In the other case, in the structure with low redundancy,  the real pruning threshold is lower than the unified pruning threshold and the significant channels may be pruned in this structure, which seriously affects the accuracy of the network. 

To solve this problem, we group the object detection network first. As shown in {Fig.3},  the blue dotted boxes represent the YOLOv4 network's backbone part, the feature enhancement part, and the detection part. The red dotted boxes represent the five groups. When input the size of 416×416 images into the network, the group1 include the scales of the feature layer from  416×416 to 52×52 and CSP module, group2 only include the feature layer at 26×26 scales along with the CSP module, group3 only include the feature layer at 13×13 scales along with the CSP and CBL module, group4 and group5 both include  the feature layer from 52×52 scales to 13×13 scales and CBL module,however, for more precise pruning, we divide this two parts into two groups. Besides, the CSP module includes the CBM module as shown in {Fig.2}. The SPP,  Concat, Upsample, and Downsample these modules are just operators, they do not contain the trainable parameters. 

The first three groups, Group1$ \sim $3 belong to the backbone network and are responsible for extracting the features of the object, these three groups contain all the residual modules in the network. Group4 belongs to the feature enhancement network, which is responsible for further enhancement and fusion of features. Group5 includes the detection head part which realizes the classification of features and the regression of object location.

\subsubsection{Pruning thread and mask matrix}
Due to the different redundancy in five groups, we need to obtain five pruning thresholds and a pruning mask matrix from five groups. Here we illustrate the steps using one group. Firstly, we calculate the ratio of the number of channels in the current group to the total number of  channels that can be pruned in the whole network, we denote this ratio as $\mathbf{p}_{i}$. Then, given a total pruning ratio of the whole network, we denote it as $P$. By multiplying $\mathbf{p}_{i}$ by $P$, we can obtain the pruning proportion of the current group, denote as $\mathbf{g}_{i}$. After that, we sort all the $\gamma$ coefficients in that group, according to the current group's pruning proportion($\mathbf{g}_{i}$) we can get the current group's pruning threshold(denote as $\mathbf{t}_{i}$).  Finally, we use the pruning threshold $\mathbf{t}_{i}$ to compare with all the $\gamma$ coefficients in this group to obtain the pruning mask matrix of the convolution layers, in that pruning mask matrix, the number 1 represents the channel in the corresponding position retained, and the number 0 represents the channel in the corresponding position will be pruned.

As shown in {Fig.4}, we use the $\gamma$ coefficient as the scale factor of the channel and compare the scale factor with the pruning threshold of the current group. When the value of the scale factor is lower than the pruning threshold, the channel corresponding to the scale factor will be pruned. Using the above method, we obtain the five pruning thresholds and the mask matrix of the convolution layer in each group.

\begin{figure}[htbp]
\begin{center}
\includegraphics[width=0.85\textwidth]{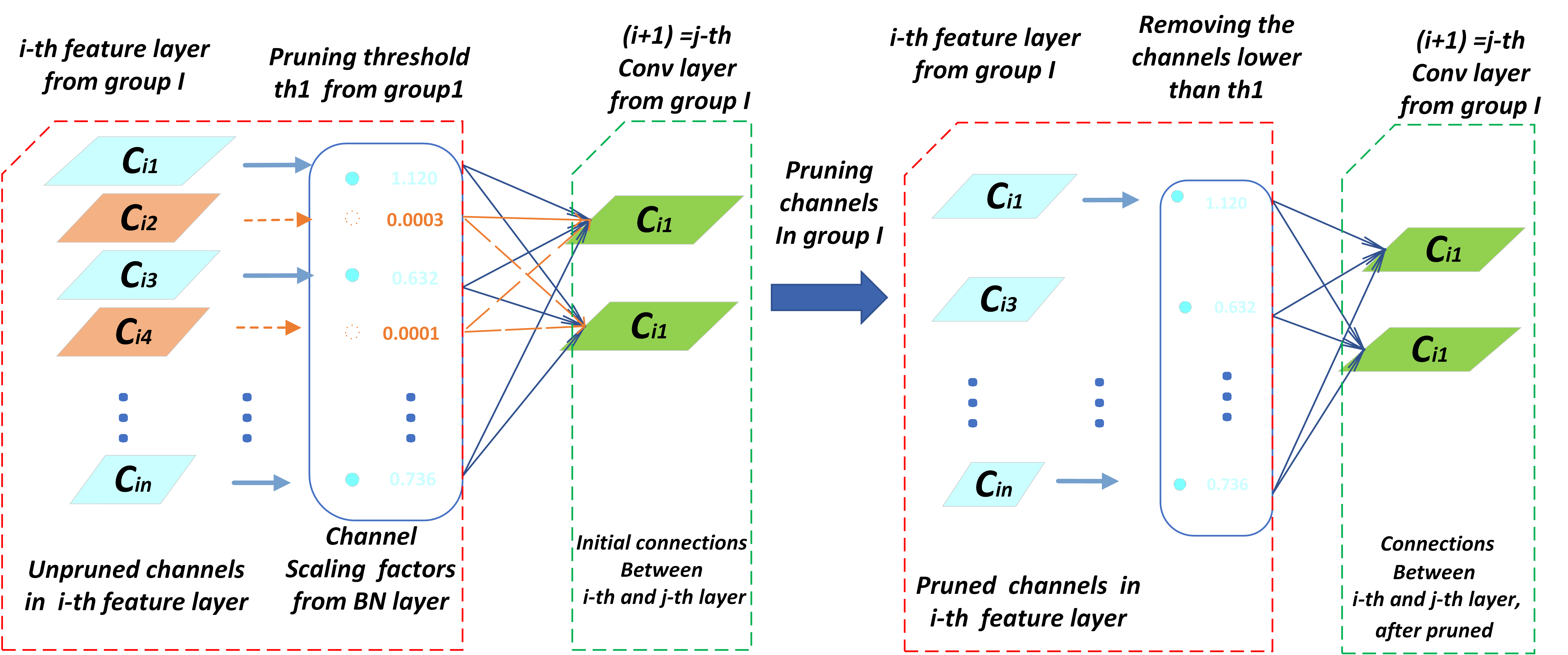}
\end{center}
\caption{The red dotted box represents that the feature layer is assigned to group I, the green dotted box represents that the convolution layer next to this feature layer. They are both in group I, and in this group, all the channels will share the same pruning threshold. We use the $\gamma$ as the scale factor of the channel. When the channel scaling factors are lower than the current group's pruning threshold, the corresponding channels will be pruned.}
\label{Fig.4}
\end{figure}

\subsubsection{Public mask matrix }
The pruning mask matrix obtained by the above method can be used as the final  pruning matrix of the most convolution layers  in the network. However, another convolution layer associated with the shortcut layer needs to use the public mask pruning matrix. Because the number of channels to be added between the two layers must be consistent in the shortcut layer to perform the addition operation. Considering that the source layer of the shortcut may still be a shortcut layer, which will involve multiple convolution layers forward, and the pruning mask matrix of these convolution layers needs to be consistent. How to generate this public mask pruning matrix so that the channel pruning in each convolution layer reaches a higher amount and has little effect on precision. This is a question worth considering. 

To address such a question, we propose a voting method to generate the public pruning mask matrix. As shown in {Fig.5}. Firstly, we count the total number of convolution layers associated with the shortcut layer and denoted as $\mathrm{N}_{\text {conv }}$. Then we count the total number of zeros at position $(i,j)$ in the pruning mask matrix and denoted as $Z_{(i,j)}$. The value of the public mask matrix at $(i,j)$ position denoted as $p_{(i, j)}$. When the ${Z_{(i, j)}}\geq\left(N_{\text {conv }} / 2\right)$,then $p_{(i, j)}=0$ ; otherwise, the $p_{(i, j)}=1$.

\begin{figure}[htbp]
\begin{center}
\includegraphics[width=0.95\textwidth]{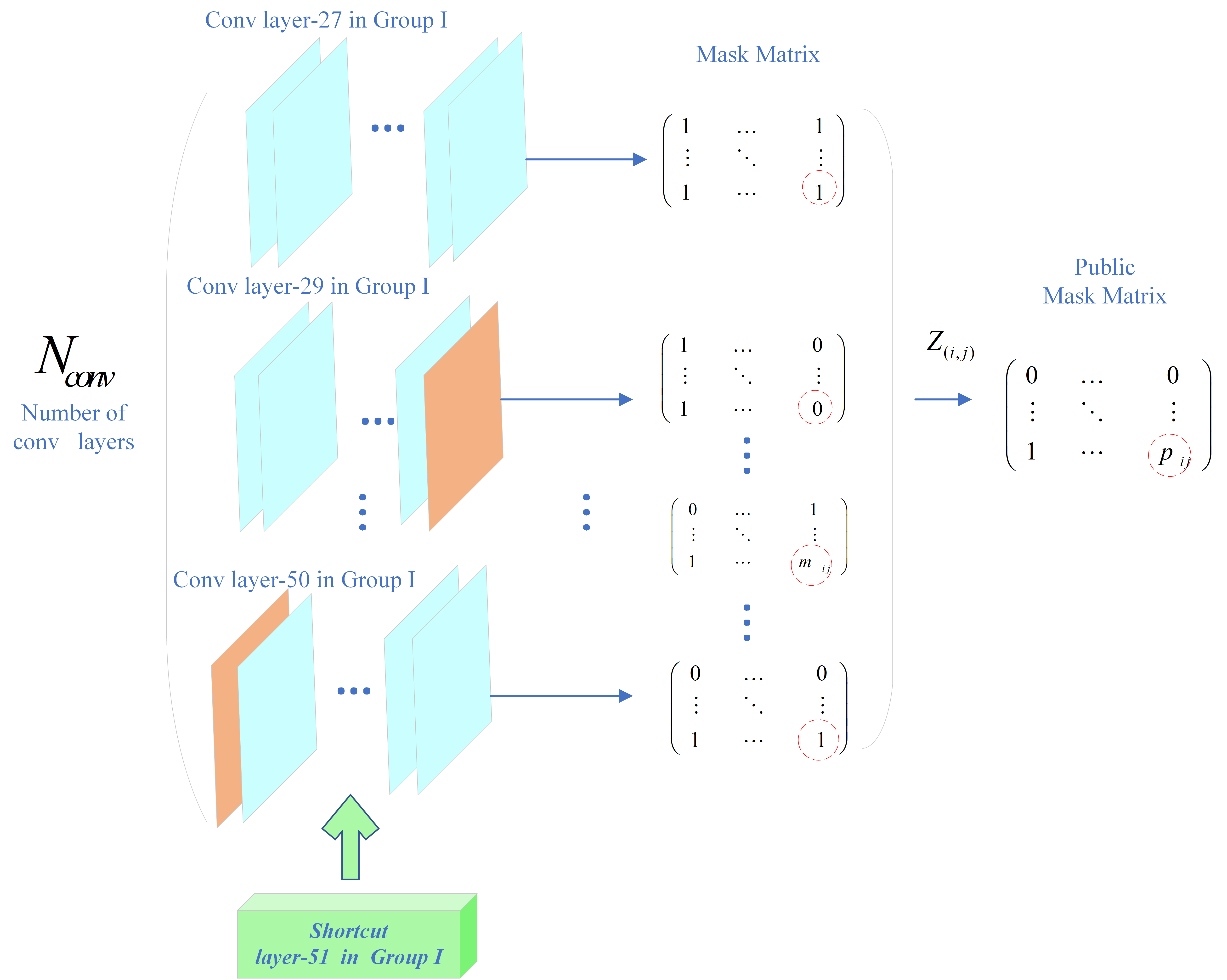}
\end{center}
\caption{Generating the public pruning mask matrix. In the figure, $\mathrm{N}_{\text {conv }}$ denotes the total number of convolution layers associated with the shortcut layer. $Z_{(i,j)}$ denotes the total number of zeros at position $(i,j)$ in the pruning mask matrix. $p_{(i, j)}$ denotes the value of the public mask matrix at $(i,j)$ position. If $Z_{(i, j)} \geq\left(N_{\text {conv }} / 2\right),$ then $\mathrm{p}_{(i, j)}=0 ;$ else, the $\mathrm{p}_{(i, j)}=1$.}
\label{Fig.5}
\end{figure}

\subsection{Knowledge distillation loss}
In this section, we introduce the three parts of distillation loss. As shown in {Fig.6}, we use the original network as the teacher network, and the pruned network as the student network for knowledge distillation, the distillation loss includes three parts: 1. The difference in spatial attention between student network and teacher network  denoted as $L_{AT}$; 2. The predicted value differences between student networks and teacher networks in object classification and location regression are denoted as $L_{\text{soft}}$; 3. The loss between the predicted value of the student network and the ground truth is denoted as $L_{\text {hard}}$.

\begin{figure*}[htbp]
\begin{center}
\includegraphics[width=0.95\textwidth]{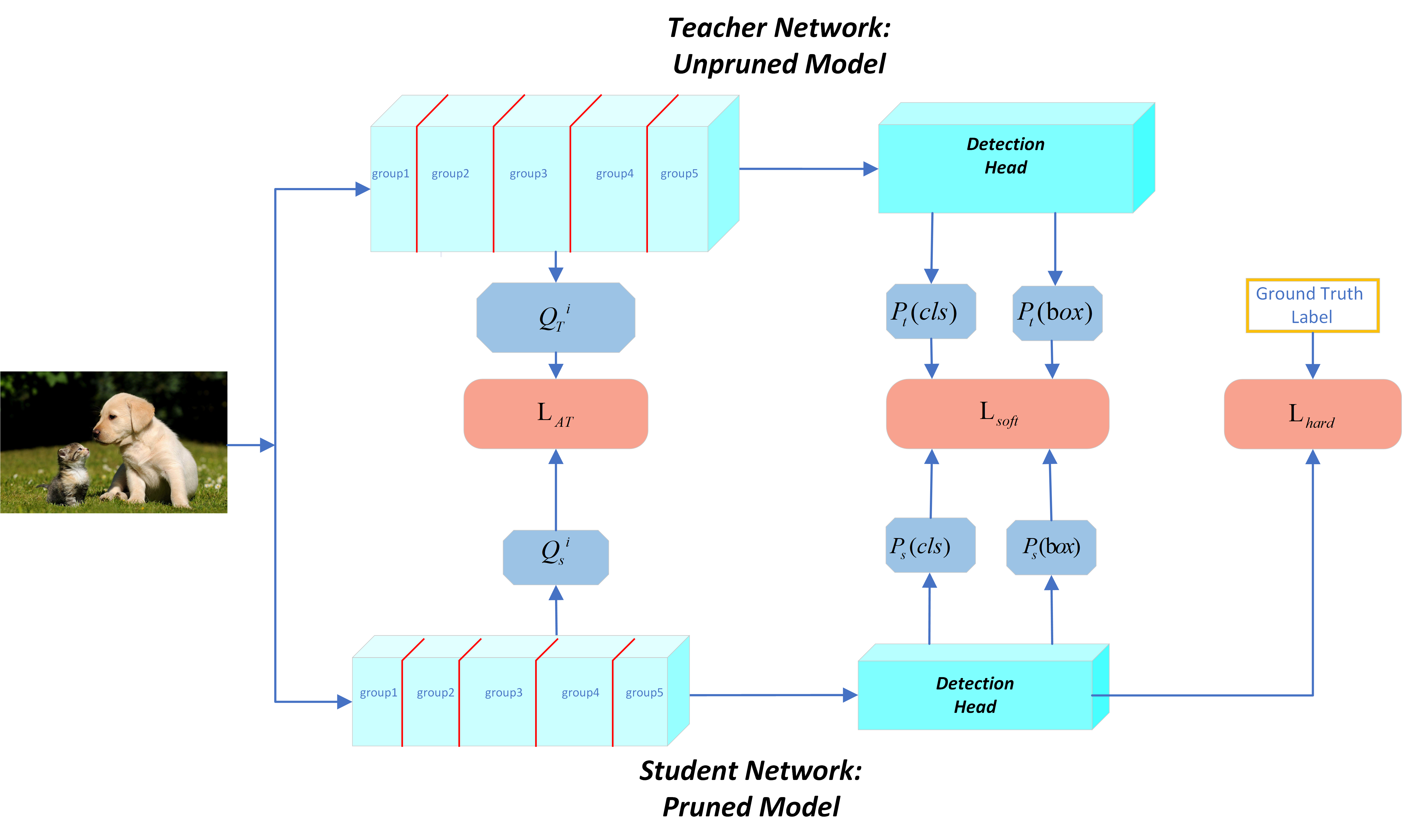}
\end{center}
\caption{Distilling for the group pruned network. $Q_{T}^{i}$ and $Q_{S}^{i}$ are the spatial attention information extract from the feature maps of five specific scales in each group of the teacher and student  network, respectively. Three red boxes demonstrated the three-loss parts of the student network. 1. $L_{AT}$ denotes the difference in spatial attention between student network and teacher network; 2.$L_{\text{soft}}$ denotes the predicted value differences between student network and teacher network in object classification and location regression; 3.$L_{\text {hard}}$ denotes the loss between the predicted value of the student network and the ground truth.}
\label{Fig.6}
\end{figure*}

As in \eqref{4}, we use $L_{\text {total }}$ to represent the total loss of the student network, and we will mainly consider the loss of $L_{AT}$ and $L_{\text{soft}}$.

\begin{equation}
L_{\text {total}}=L_{AT}+L_{\text{soft}}+L_{\text {hard}},\label{4}
\end{equation}

\subsubsection{ Group spatial attention loss}
As in \eqref{5}, $L_{AT}$ denotes the difference in spatial attention information between student network and teacher network. We reduce this difference by allowing the student network to imitate the spatial attention of the teacher network. 

\begin{equation}
\mathrm{L}_{A T}=\sum_{i=1}^{5} \beta_{i}\left\|\frac{Q_{T}^{i}}{\left\|Q_{T}^{i}\right\|_{2}^{i}}-\frac{Q_{S}^{i}}{\left\|Q_{S}^{i}\right\|_{2}}\right\|_{2},\label{5}
\end{equation}

where $i$ belongs to 1$ \sim $ 5, representing the five groups in the network. From these five groups, we extract the spatial attention only at specific scale feature maps as the knowledge, the scale feature maps are 208 × 208, 104 × 104, 52 × 52, 26 × 26 and 13 × 13, respectively,  We extract these feature maps from five groups. $\beta_{i}$ denotes the five group's loss gain coefficient, we give the different group's spatial attention with different weight. $Q_{T}^{i}$ and $Q_{S}^{i}$ are the 1-dimensional tensor forms of teacher and student network spatial attention, and each element in $Q^{i}$ is normalized.

As in \eqref{6}, one-dimensional $Q^{i}$ is converted from the two-dimensional $F\left(A^{i}\right)$ by the flattening operation. $F\left(A_{T}^{i}\right)$ and $F\left(A_{S}^{i}\right)$ are two-dimensional matrix forms of spatial attention in the network of teacher and students, respectively. 

\begin{equation}
Q_{\mathrm{T}}^{i}=\operatorname{vec}\left(F\left(A_{T}^{i}\right)\right), Q_{S}^{i}=\operatorname{vec}\left(F\left(A_{S}^{i}\right)\right), \label{6}   
\end{equation}

The mapping function $F(.)$ is given in \eqref{7}, where $A$ denotes the feature map on the channel, $A$ has H × W size, and $C$ represents the number of all channels on the feature layer. The value of $p$ is 2, which represents a power of 2 for each element in $A$.

\begin{equation}
F(A)=F_{s u m}^{p}(A)=\sum_{j=1}^{c}\left|A_{j}\right|^{p}, \label{7}  
\end{equation}

Spatial attention refers to extracting the spatial information of all channels on a certain feature layer in the form of a heat map. The extraction process is shown in {Fig.7}. One feature layer is selected from the network, and the size of the feature layer is C × H × W.  C represents the number of channels on the feature layer. Through the mapping function F: $F: \mathrm{A}^{\mathrm{C} \times H \times W} \rightarrow \mathrm{A}^{H \times \mathrm{W}}$, the 3-dimensional feature layer tensor is mapped to a 2-dimensional tensor on the channel dimension, which represents the spatial attention map of the feature layer.

\begin{figure}[htbp]
\begin{center}
\includegraphics[width=0.5\textwidth]{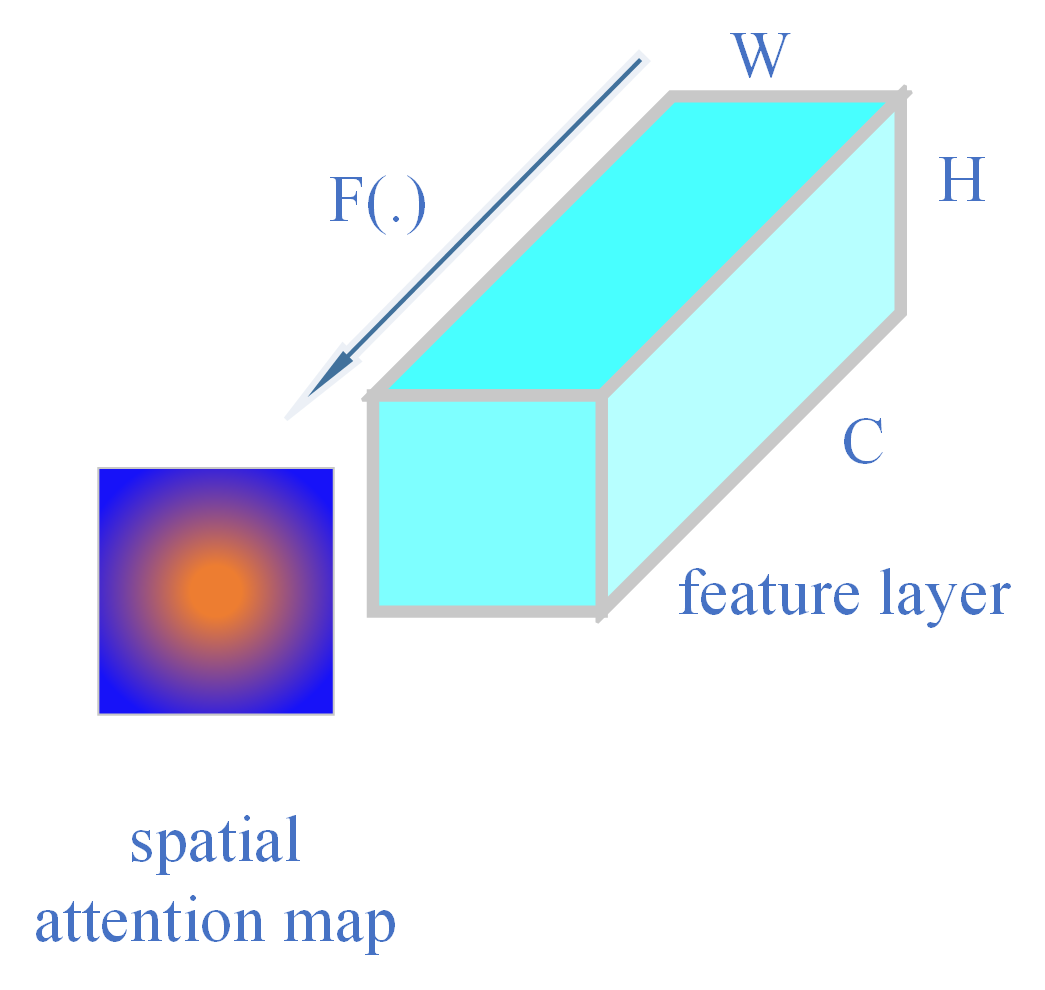}
\end{center}
\caption{Generating the spatial attention map from the feature layer.}
\label{Fig.7}
\end{figure}

\subsubsection{Soft target loss }
As in \eqref{8}, $L_{\text{soft}}$ is composed of two kinds of prediction differences. One is the prediction difference of teacher and student networks in object classification. The other is the prediction difference of teacher and student networks in the location and size of the object box.  
\begin{equation}
L_{\text{soft}}=l_{(t-s)}(cls)+l_{(t-s)}(box),\label{8}
\end{equation}

$l_{(t-s)}(cls)$ denotes the prediction difference in object classification between teacher networks and student networks, as shown in \eqref{9}.
\begin{equation}
l_{(t-s)}(cls)= \frac{1}{k} \sum_{i=1}^{3} \sum_{j=1}^{k}M^{(i, j)} \left(\log M^{(i, j)}-\mathrm{N}^{(i, j)}\right),\label{9}
\end{equation} 

where $i$ denotes the prediction of the network at three scales, $k$ denotes the number of all prior boxes at the current scale. $M^{(i, j)}$ denotes the predicted output of the teacher network after distillation. ${N}^{(i, j)}$ denotes the predicted output of the student network after distillation. 

As in \eqref{10} \eqref{11}, $M^{(i, j)}$, ${N}^{(i, j)}$ is obtained by $softmax$ and  $logsoftmax$ function, where $P_{t}^{(i, j)}(cls)$ denotes the classification probability predicted for each prior box in the teacher network and $P_{s}^{(i, j)}(cls)$ denotes the classification probability predicted for each prior box in the student network. $T$ is a temperature parameter used to make the output distribution of teacher and student network prediction more uniform.

\begin{equation}
\begin{aligned}
M^{(i, j)} & =\operatorname{soft} \max \left(P_{t}^{(i, j)}(cls)/ T\right),\label{10}
\end{aligned}
\end{equation}

\begin{equation}
N^{(i, j)} =\log _{-} \operatorname{soft} \max \left(P_{s}^{(i, j)}(cls) / T\right),\label{11}
\end{equation} 

As in \eqref{12}, $l_{(t-s)}(box)$ denotes the prediction difference between teacher network and student network on the location and size of the object box.

\begin{equation}
\begin{aligned}
& l_{(t-s)}(box)=\sum_{i=1}^{3} \sum_{j=1}^{k}\left\|P_{t}^{(i, j)}(box)-P_{s}^{(i, j)}(box)\right\|_{2}, 
\label{12}
\end{aligned}
\end{equation}

where $i$ denotes the prediction of the network at three scales, and $k$ denotes the number of remaining candidate boxes after meeting the IOU thread at the current scale. In the position corresponding to the object candidate box, in the student network the position and size of the predicted box was denoted by the $P_{s}^{(i, j)}(box)$. In the teacher network, the position and size of the predicted box was denoted by the $P_{t}^{(i, j)}(box)$.

\section{Experiments}
In the experiment, we take the YOLOv4 detection network and PASCAL VOC data sets as an example to illustrate and validate the effectiveness of the proposed model compression. Firstly, we sparsely train the network with a dynamic sparse rate. Secondly, we quantitatively analyze the effect of different pruning proportions on the model size, accuracy, and calculation. Then, we compare our group pruning with other current pruning methods on object detection data sets. At last, to prove the superiority of the distillation method, we compare the accuracy of the pruned network after fine-tuning and distilling. Furthermore, we combine our distillation method with other common pruning methods to demonstrate that our distillation method was suitable for other common pruning methods.

\subsection{Dataset and evaluation metrics}
For the dataset, we use Pascal VOC \cite{everingham2015pascal}. The voc2012train, voc2012val, voc2007train, and voc2007val, these four parts have 16551 pictures and we combine them used as the final training set. The voc2007test included 4952 pictures and was used as the final test set. Our experimental environment is Ubuntu 18.04, PyTorch = 1.8 version, GPU is a single RTX3090. For the evaluation metrics, we evaluate the performance of the pruned model from four aspects: model size, the number of parameters, calculation amount, and mAP@0.5. The computation is measured by FLOPS. The mAP@0.5 represents the average value of all categories of AP when the IOU threshold is 0.5, AP represents the average precision of one category, and the specific calculation details are referred to in reference \cite{padilla2020survey}.

\subsection{Dynamic Sparse training}
In the process of sparse training, the L1 norm of $\gamma$ is added into the loss function as a regularization term to train together. As shown in {Fig.8}, the distribution of $\gamma$ coefficients of all BN layers in the original network is in different ranges. The sparse training process is to make the $\gamma$ coefficient distribution close to zero, so that is convenient to select out the unimportant channels in the network. 

\begin{figure}[htbp]
\begin{center}
\includegraphics[width=0.8\textwidth]{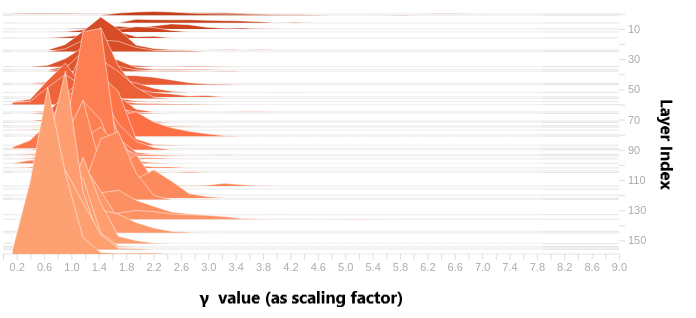}
\end{center}
\caption{The distribution of $\gamma$ coefficients in the original network.}
\label{Fig.8}
\end{figure}

During the experiment, we found that sparse training is the trade-off between accuracy and sparsity. A larger sparse rate $s$ can bring a better sparse effect, but the accuracy loss is also large, even if the number of epochs of sparse training is increased in the future, the model still can not restore to good accuracy. A smaller sparse rate $s$ has little effect on the accuracy but leads to a worse sparse effect. To solve this problem and make a good balance between sparse effect and accuracy, we put forward a dynamic sparse training method, which introduces a variable sparse rate $s$, and the sparse rate $s$ will change with the training process of the network. 

During the dynamic sparse training, the degree of network sparsity can be adjusted by setting the sparsity rate s, the network will dynamically adjust its sparse rate according to the number of epochs of current training. For the YOLOv4 network, we set the initial sparse rate $s = 0.00075$, the initial learning rate $lr0 = 0.002$ and train 200 epochs. When training to half the number of epochs, 70 $\%$ of the channel maintains the original sparse rate, 30 $\%$ of the channel sparse rate decay to 1 $\%$ of the original sparse rate. Besides, the learning rate is updated by cosine annealing. When the input image size is 416×416, the batch size is set to 16, as shown in {Fig.9}, this figure represents the $\gamma$ coefficient distribution in the network layers after dynamic sparse training, compared to the original the $\gamma$ coefficient distribution (as shown in {Fig.8}), most of these $\gamma$ coefficient is more close to the zero and it is convenient to select out insignificant channels.

\begin{figure*}[h]
\begin{center}
\includegraphics[width=0.8\textwidth]{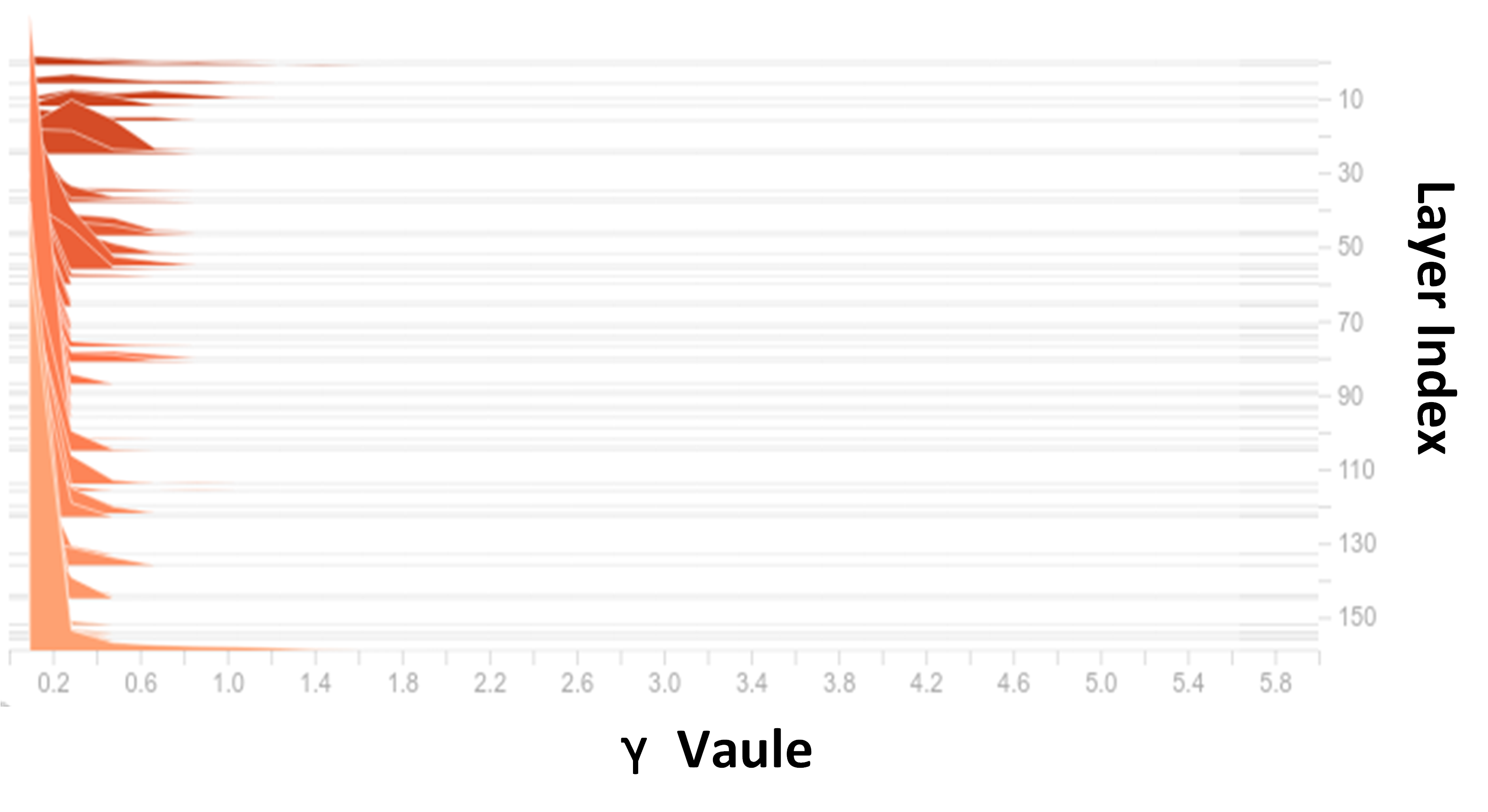}
\hspace{15mm}
\end{center}
\caption{ The distribution of the $\gamma$ coefficient in the network after dynamic sparse training.}
\label{Fig.9}
\end{figure*}

\subsection{Group channel pruning}

In this section, we divide the network layers into five groups. YOLOv4 has 162 layers, during the group, we comprehensively consider the scale of feature layers and the similarity of module structure in the network which means that the feature layers with the same scale and the layers that have similar module structure are assigned to the same one group.

When input the size of 416×416 images into the network, the group1 include the scales of the feature layer from  416×416 to 52×52 and the CSP module, group2 only include the feature layer at 26×26 scales along with the CSP module, group3 only include the feature layer at 13×13 scales along with the CSP and CBL module, group4 and group5 both include the feature layer from 52×52 scales to 13×13 scales and CBL module. The specific layers in each group are the following, including Group1: 0 $ \sim $ 55 layers, Group2:56 $ \sim $ 85 layers, Group3: 86 $ \sim $ 116 layers, Group4:117 $ \sim $ 136 layers, and Group5:136 $ \sim $ 161 layers.

\begin{table*}[htbp]
\flushleft
\captionsetup{justification=raggedright}
\caption{ The Total represents the total pruning proportion of the whole detection network, the Model represents the compressed model under the corresponding total pruning proportion  and the  Group1$ \sim $ 5 shows the specific pruning proportion in 5 groups.}

\setlength{\tabcolsep}{4mm}{
\begin{tabular}{lccccccccc} 
\toprule
 Total &  Model  & Group1  &  Group2   &  Group3   &  Group4   &  Group5 &  Model Size &   mAP@0.5 \\
\midrule
  0    &     Base    &  0    & 0     & 0    & 0     & 0    & 256M & 79.8  \\
  38$\%$  & Model1  &  10$\%$ & 20$\%$  & 96$\%$ & 87$\%$  & 45$\%$ & 98MB  & 73.6  \\
  40$\%$  & Model2  &  10$\%$ & 25$\%$  & 96$\%$ & 87$\%$  & 50$\%$ & 90MB  & 56.3  \\
  42$\%$  & Model3  &  10$\%$ & 25$\%$  & 97$\%$ & 85$\%$  & 55$\%$ & 84MB  & 28.3  \\
  45$\%$  & Model4  &  15$\%$ & 25$\%$  & 96$\%$ & 85$\%$  & 70$\%$ & 77MB  & 9.2   \\
  50$\%$  & Model5  &  15$\%$ & 27$\%$  & 96$\%$ & 85$\%$  & 90$\%$ & 68MB  & 6.1   \\
\bottomrule
\end{tabular}

    }

\flushleft
  \label{Table1}
\end{table*}

\subsubsection{Reducing model's parameters and computations}

Given a total pruning proportion of the whole network, according to the total pruning proportion the algorithm will calculate the five groups' pruning proportion. As shown in {Table1}, we demonstrate the details of five groups' pruning proportion. It can be seen from {Table1} that in the backbone network part, the function of feature extraction is mainly realized by Group1 and Group2, and the redundancy in these two groups reaches 10$\%$ $\sim$  25 $\%$. The redundancy in Group3 reached more than 90$\%$, this indicating that the channels in these feature layers have little effect on feature extraction in Group3. The redundancy in Group4 reaches more than 90$\%$, and most channels in this Group play a little effect on feature enhancement. The redundancy in Group5 reaches about 45$\%$.
 
Through the above analysis, which is sufficient to indicate that the redundancy in various structural parts of the network is different. We use group pruning to make each group has different pruning thresholds, thus achieving more accurate and efficient pruning. In {Table2}, we present the effects of different pruning proportions on the model's parameters and computation, except the pruning proportion is different, we keep the size of input images is 416 × 416.  {Table2} demonstrates the effectiveness of our group channel pruning.

\begin{table}[htbp]
\centering
\caption{Comparing to the original network, the reduction in model parameters and computations by using our group pruning method with different pruning proportions. Table2 $ \sim $ Table 5, for all models, the size of the input image is 416 × 416. }
\setlength{\tabcolsep}{0.5mm}{
\begin{tabular}{lccccc} 
\toprule
  Model  & Model Size  &  Flops   &  Pruned  & Parameters  &  Pruned   \\
  \midrule
  Base    &  256MB   & 29.9G      & 0           & 64.1M     & 0         \\
  Model1  &  98MB    & 20.35G    & 31.9$\%$    & 24.51M    & 61.7$\%$   \\
  Model2  &  90MB    & 19.46G    & 34.9$\%$    & 22.56M    & 64.7$\%$   \\
  Model3  &  84MB    & 19.35G    & 35.2$\%$    & 20.96M    & 67.2$\%$   \\
  Model4  &  77MB    & 17.94G    & 40.0$\%$    & 19.27M    & 69.9$\%$   \\
  Model5  &  68MB    & 16.57G    & 44.6$\%$    & 17.07M    & 73.3$\%$   \\
\bottomrule
\end{tabular}
}
\label{Table 2}
\end{table}

\subsubsection{Comparing with other current pruning methods}
As shown in {Fig.10}, to demonstrate the superiority of our proposed group channel pruning method in object detection, we quantitatively compare our method with other current pruning methods like Network Slimming \cite{liu2017learning}, Thinet \cite{luo2017thinet}, Layer pruning \cite{jordao2020discriminative} and  Eagle eye \cite{li2020eagleeye}. We compare them at two aspects: model size and accuracy (map@0.5).

\begin{figure*}[htp]
\begin{center}
\includegraphics[width=0.8\textwidth]{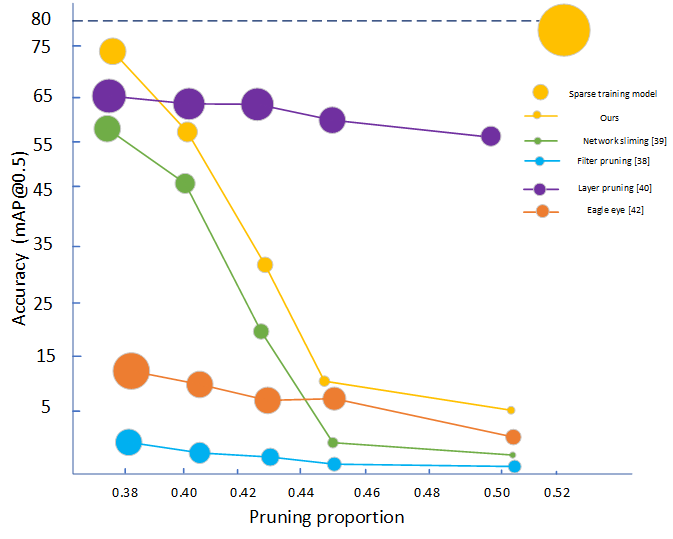}
\hspace{30mm}

\end{center}
\caption{The figure shows the accuracy of the object detection network changes with the pruning ratio, and the  dot radius represent the size of the  model changes with the pruning ratio, the input resolution is 416 × 416.}
\label{Fig.10}
\end{figure*}

During the pruning process, we keep the pruning proportion as the same for different pruning methods, the size of input images is 416 × 416. It can be seen from {Fig.10}, under the same pruning proportion, our method can obtain the best trade-off between the pruned model's accuracy and pruned model's size.

In addition, in \cite{jordao2020discriminative} they use a subspace projection approach to estimate the importance of the network layers, when using this  pruning way, the layers can be pruned is limited since that if the pruning proportion is above 0.6 this pruning way will change the network architecture and accuracy significantly, this is inconvenient to recover the pruned models' accuracy.In \cite{li2020eagleeye}, they  use a way similar to the  network architecture search, during the pruning process,  they search the pruned network not only consider the pruned model's size, accuracy, but also consider the pruned model's computations and selected the best trade-off model from 1000 candidate models. To ensure  the  work \cite{li2020eagleeye} pruning method is carried out under the same hardware conditions and the computational environment with other pruning methods (Our experimental environment is Ubuntu 18.04, PyTorch = 1.8 version, GPU is a single RTX3090), we choose the best model only from the five candidate pruning models, we also need to declare that this pruning method of architecture search has better performance when the number of candidate models is more, but this situation also puts forward higher requirements for computational power.

\subsection{Group spatial attention distilling}
In this section, we use the sparse network to conduct the group channel pruning and obtain the pruned network. The accuracy of the original network and the network after sparse training are 87.1 and 79.8, respectively.

During the distillation experiments, we set the original network as the teacher network and  the pruned network as the student network, the spatial attention information was extracted only at specific scale feature maps from five groups as the knowledge, the scale of feature maps are 208 × 208, 104 × 104, 52 × 52, 26 × 26 and 13 × 13, respectively.   Besides, we give the five group's loss gain coefficient $\beta_{i}$ with different weight,  we set the $\beta_{1}$- $\beta_{3}$ as 1000 and the $\beta_{4}$- $\beta_{5}$  as 10000.

\subsubsection{Group  spatial attention distilling with our  pruning method}
To verify the effectiveness of group spatial attention distilling, we use it for the pruned model with our pruning method. Then, we fine-tune and group attention distills the compressed model, respectively.

The comparison results were shown in {Table 3}, the mAP@0.5 as the accuracy evaluation metric. It can be seen from {Table 3 } when directly fine-tune the pruned network, the highest accuracy restored can only reach 81.1. In contrast, using group spatial attention distilling  can make  the pruned network obtain higher accuracy. 

\begin{table}[htbp]
\centering
\caption{Combing the group spatial attention distilling with our pruning method, the pruned Model1-5 achieved by ours' pruning method, the accuracy represent the Model after through the fine-tuning and group spatial attention (denote as GSA) distillation.}
\begin{tabular}{llccccc} 
\toprule
\multirow{2}{*}{Total} & \multirow{2}{*}{Model} & \multirow{2}{*}{Model Size }
& \multicolumn{2}{c}{mAP@0.5}  \\
\cline{4-5}  
&&& Fine tuning &  GSA Distilling \\
  \midrule
  38$\%$  & Model1  &  98MB    & 80.4    & 86.5     \\
  40$\%$  & Model2  &  90MB    & 81.1    & 86.6     \\
  42$\%$  & Model3  &  84MB    & 80.5    & 84.8     \\
  45$\%$  &  Model4  &  77MB    & 79.5    & 83.3     \\
  50$\%$  &   Model5  &  68MB    & 75.9    & 76.3     \\

  \bottomrule
\end{tabular}
\label{Table 3}
\end{table}

\subsubsection{Group  spatial attention distilling combine with other common pruning methods}

To show that our group spatial attention distillation scheme is not only suited for our pruning method, we  combine the group spatial attention distillation  method  with other common pruning methods like Network Slimming \cite{liu2017learning} and Thinet \cite{luo2017thinet}.

The comparison results were shown in {Table 4 $ \sim $ 5}, it can be seen from {Table 4 $ \sim $ 5 }, when we directly fine-tune the pruned network, the highest accuracy restored can only reach 80.9. In contrast, using group spatial attention distilling  can make  the pruned network obtain higher accuracy. Besides, combing the group channel pruning with our pruning method can achieve a better effect.

\begin{table}[htbp]
\centering
\caption{Combing the group spatial attention (denote as GSA) distilling with Network Sliming \cite{liu2017learning} pruning method, the pruned Model1-5 achieved by \cite{liu2017learning} .}
\begin{tabular}{llccccc} 
\toprule
\multirow{2}{*}{Total} & \multirow{2}{*}{Model} & \multirow{2}{*}{Model Size }
& \multicolumn{2}{c}{mAP@0.5}  \\
\cline{4-5}  
&&& Fine tuning &  GSA Distilling \\
  \midrule
  38$\%$  & Model1  &  100MB  &  80.4   &   83.1   \\
  40$\%$  & Model2  &  94MB   & 79.8    & 83.7    \\
  42$\%$  & Model3  &  86MB    &  80.4   & 84.8    \\
  45$\%$  &  Model4  &  78MB    & 79.6    &  82.8  \\
  50$\%$  &   Model5  &  65MB    & 79.3    & 80.9  \\

  \bottomrule
\end{tabular}
\label{Table 4}
\end{table}

\begin{table}[htbp]
\centering
\caption{Combing the group spatial attention distilling (denote as GSA) with Thinet\cite{luo2017thinet} pruning method, the pruned Model1-5 achieved by \cite{luo2017thinet}.}
\begin{tabular}{llccccc} 
\toprule
\multirow{2}{*}{Total} & \multirow{2}{*}{Model} & \multirow{2}{*}{Model Size }
& \multicolumn{2}{c}{mAP@0.5}  \\
\cline{4-5}  
&&& Fine tuning &  GSA Distilling \\
  \midrule
  38$\%$  & Model1  &  116MB    & 80.9    & 84.2    \\
  40$\%$  & Model2  &  110MB    & 80.4    & 83.9     \\
  42$\%$  & Model3  &  104MB    & 80.9    & 81.8     \\
  45$\%$  &  Model4  &  96MB    & 80.7    & 82.1     \\
  50$\%$  &   Model5  &  83MB    & 79.4    & 82.1     \\

  \bottomrule
\end{tabular}
\label{Table 5}
\end{table}

\subsubsection{Comparing the compressed model with other object detection network on PASCAL and COCO}
To verify the effectiveness of our final compressed model, we compare the final compressed model with other normal network and lightweight detectors  on PASCAL and COCO, respectively.

\begin{table}[htbp]
\centering
\caption{ Comparing with normal and lightweight detectors on PASCAL VOC. }
\setlength{\tabcolsep}{0.8mm}{
\begin{tabular}{lccccc} 
\toprule

Method  & Backbone  &   Input size  &  Flops & Parameters  &  mAP  \\
\midrule
SSD Lite \cite{sandler2019mobilenetv2}  &  VGG16        & 512×512  & 99.5B & 36.1M  & 80.7   \\
SSD Lite \cite{sandler2019mobilenetv2} &  MobilenetV2  & 320×320  & 0.8B  & 4.3M   & 71.8   \\
Tiny-DSOD \cite{litiny} & G/32-48-64-801 & 300×300 & 1.06B & 0.95M  & 72.1   \\
ThunderNet \cite{qin2019thundernet} & SNet146      & 416×416 & 0.461B & ---   & 73.8   \\
Pelee \cite{wang2018pelee}  &  PeleeNet   & 304×304   & 1.21B      & 5.43M & 70.9   \\
\midrule
Ours &  CSPDarknet*   & 320×320   & 18.2B    & 20.9M    & 82.7   \\
Ours &  CSPDarknet*   & 416×416   & 19.3B    & 20.9M    & 84.8   \\  
\bottomrule
\end{tabular}
}
\label{Table 6}
\end{table}

\begin{table}[htbp]
\centering
\caption{Comparing with normal and lightweight detectors on on COCO.}
\setlength{\tabcolsep}{0.8mm}{
\begin{tabular}{lcccccc} 
\toprule

Method  & Backbone  &  Input size  & Flops & Parameters  & mAP(0.5:0.95) &  mAP(0.5) \\
\midrule
SSD \cite{liu2016ssd}    &  VGG16      & 300×300 & 70.4B   & 34.3M & 25.7 & 43.9 \\
YOLOV3 \cite{farhadi2018yolov3} &  Darknet53  & 416×416 & 65.9B   & 62.3M & 31   & 55.3 \\
PANet \cite{liupath}& CSPResNeXt50 & 416×416 & 47.1B   & 56.9M & 36.6 & 58.1 \\
SSD lite \cite{sandler2019mobilenetv2}& Mobilenet & 320×320 & 0.8B    & 4.3M  & 22.1 & -- \\
ThunderNet \cite{qin2019thundernet}& SNet146 & 320×320 & 0.95B   & ---   & 23.6 & 40.2 \\
Pelee \cite{wang2018pelee} &  PeleeNet   & 304×304 & 2.58B   & 5.98M & 22.4 & 38.3\\
\midrule
Ours &  CSPDarknet*   & 320×320   & 18.6B    & 23.63M    & 30.2 & 48.8  \\
Ours &  CSPDarknet*   & 416×416   & 19.43B    & 23.63M    & 33.4  & 53.5 \\  
\bottomrule
\end{tabular}
}
\label{Table 7}
\end{table}

The comparison results were shown in {Table 6} and {Table7}, the mAP@0.5 and mAP@0.5:0.95 as the accuracy evaluation metric for the PASCAL VOC and COCO dataset, respectively. The symbol * represent the final compressed model which has been used our  group channel pruning, spatial attention distillation. It can be seen from above table, our method can achieve a best trade off between the network's  accuracy and calculation or parameters.

\subsection{Deployment on the edge device}
In this section,  we introduce the  deployment of the pruned model on edge device-Jetson Nano. Jetson Nano is a small, powerful computer for embedded applications, it has 128 NVIDIA CUDA cores and 4 GB  memories. We deployed the original network and five compressed models (using our  group channel pruning method) on this edge device, and test the inference time  of each model.

The specific deployment steps are as follows: Firstly, on the host machine, we prepare the network model file and the corresponding weight file, then using PyTorch to convert it to the ONNX format model. After that, on the target device-Jetson Nano, we use TensorRT to generate the engine files according to the ONNX model. At last, we run the engine files of five compression models on Jetson Nano, and the inference results are shown in {Table 6}.

It can be seen from {Table 8} that the inference time of the original network model needs 414ms and the compressed model(68M) only needs 274ms. The above experiments show that the proposed group channel pruning method can be deployed to the edge device without  designing special hardware or software and has an acceleration effect.

\begin{table}[htbp]
\centering
\caption{The inference time of the pruned model  on edge device-Jeston Nano, the input resolution is 416 × 416.}
\setlength{\tabcolsep}{0.9mm}{
\begin{tabular}{lccccc} 
\toprule
  Model  & Model Size  &  Inference time     \\
  \midrule
   Base    &  256MB   & 414 ms             \\
   Model1  &  98MB    & 311 ms    \\
   Model2  &  90MB    & 305 ms       \\
   Model3  &  84MB    & 303 ms       \\
   Model4  &  77MB    & 291 ms    \\
   Model5  &  68MB    & 274 ms     \\
   \bottomrule
\end{tabular}
}
\label{Table 8}
\end{table}

\section{Ablation Studies}
To demonstrate the generality and effectiveness of our method.  In this section, we first use MobileNet, DarkNet53, and CSPDarknet as the backbone to construct the detection network. Next, we introduce the ablation experiments of dynamic sparse training, group channel pruning, and  spatial attention distilling. At last, we test the pruning model after distillation.

\subsection{Ablation Studies for Dynamic Sparse Training}

During the sparse training, we set the total number of epochs for the sparse training and dynamic sparse training both are 200 epochs, all model's initial learning rate set as $lr0 = 0.002$ and the size of input images is 416 × 416. 

As shown in {Fig.11}, the top three pictures represent the common sparse training and  the bottom three pictures represent dynamic sparse training for the CSPDarknet-Yolov4. In the dynamic sparse training process, the batch size =16, the initial sparse rate $s = 0.00075$,  when it comes to 40 $\%$ of the total number epochs, 70 $\%$ of the channel maintains the initial sparse rate $s $, 30 $\%$ of the channel sparse rate decay to 1 $\%$ of the initial sparse rate $s$. It can be seen from {Fig.11},  the accuracy of the network through the common sparse training is 71.2, while the accuracy of the network through the dynamic sparse training is 79.8. Besides, the initial sparse rate $s$ for DarkNet53-Yolov3 and MobileNet-Yolov3 are 0.003 and 0.005, respectively. The batch size for these two networks are both 32.

\begin{figure}[t]
\centering  
\subfigure[Image 1]{
\label{Fig.sub.a}
\includegraphics[width=6.2cm,height = 3.8cm]{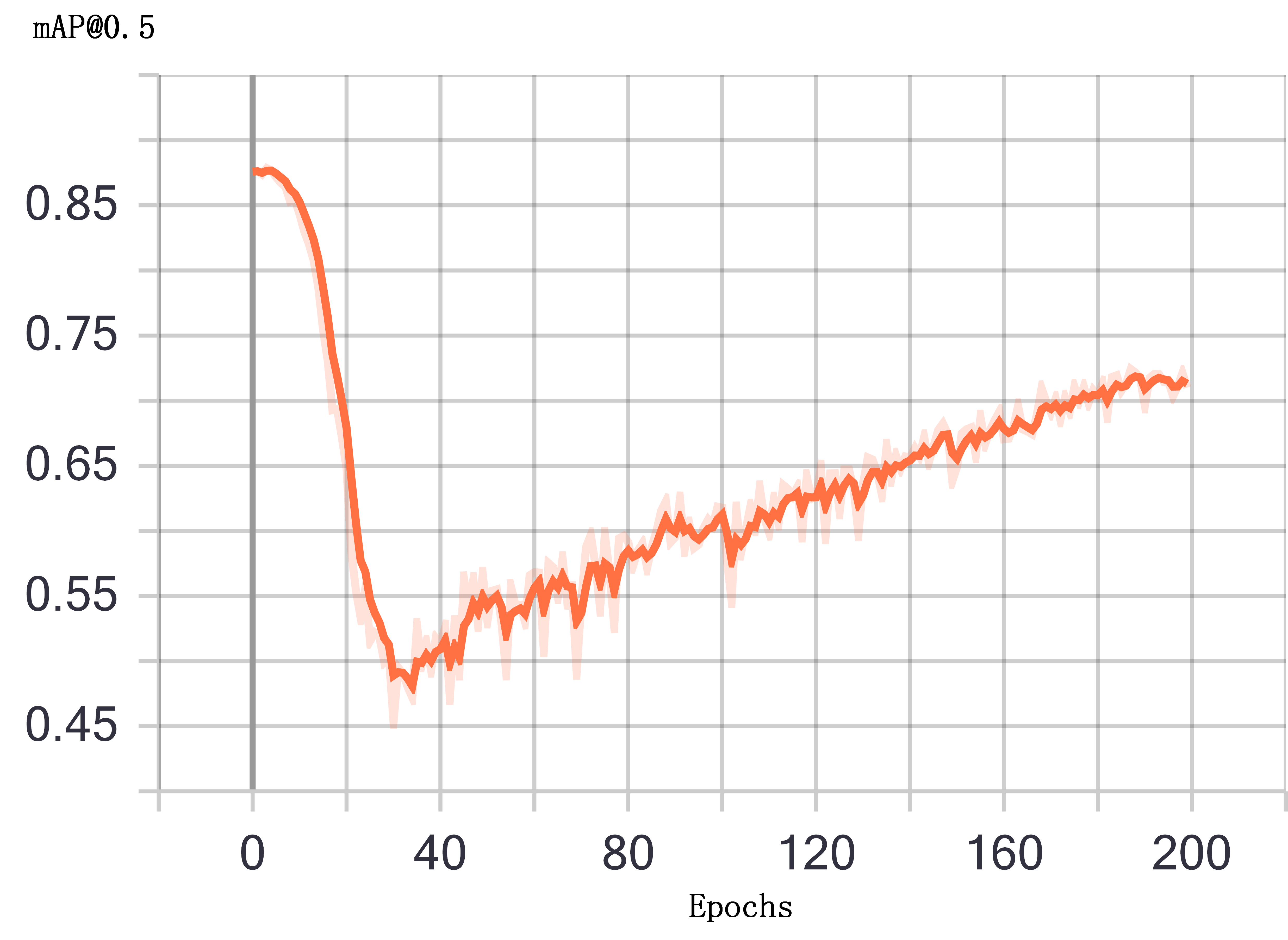}}\subfigure[Image 2]{
\label{Fig.sub.b}
\includegraphics[width=6.2cm,height = 3.8cm]{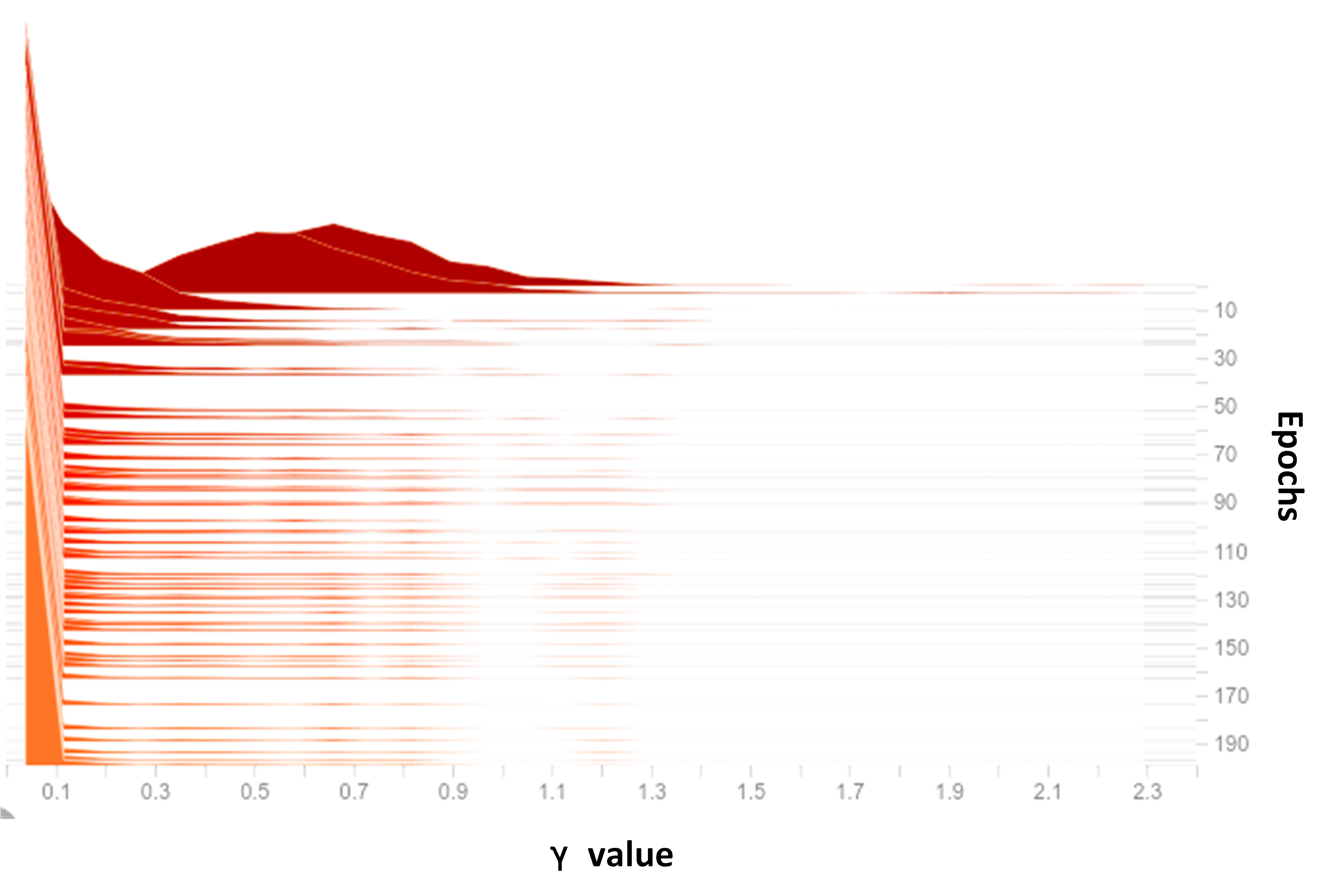}}

\subfigure[image 3 ]{
\label{Fig.sub.c}
\includegraphics[width=6.2cm,height = 3.8cm]{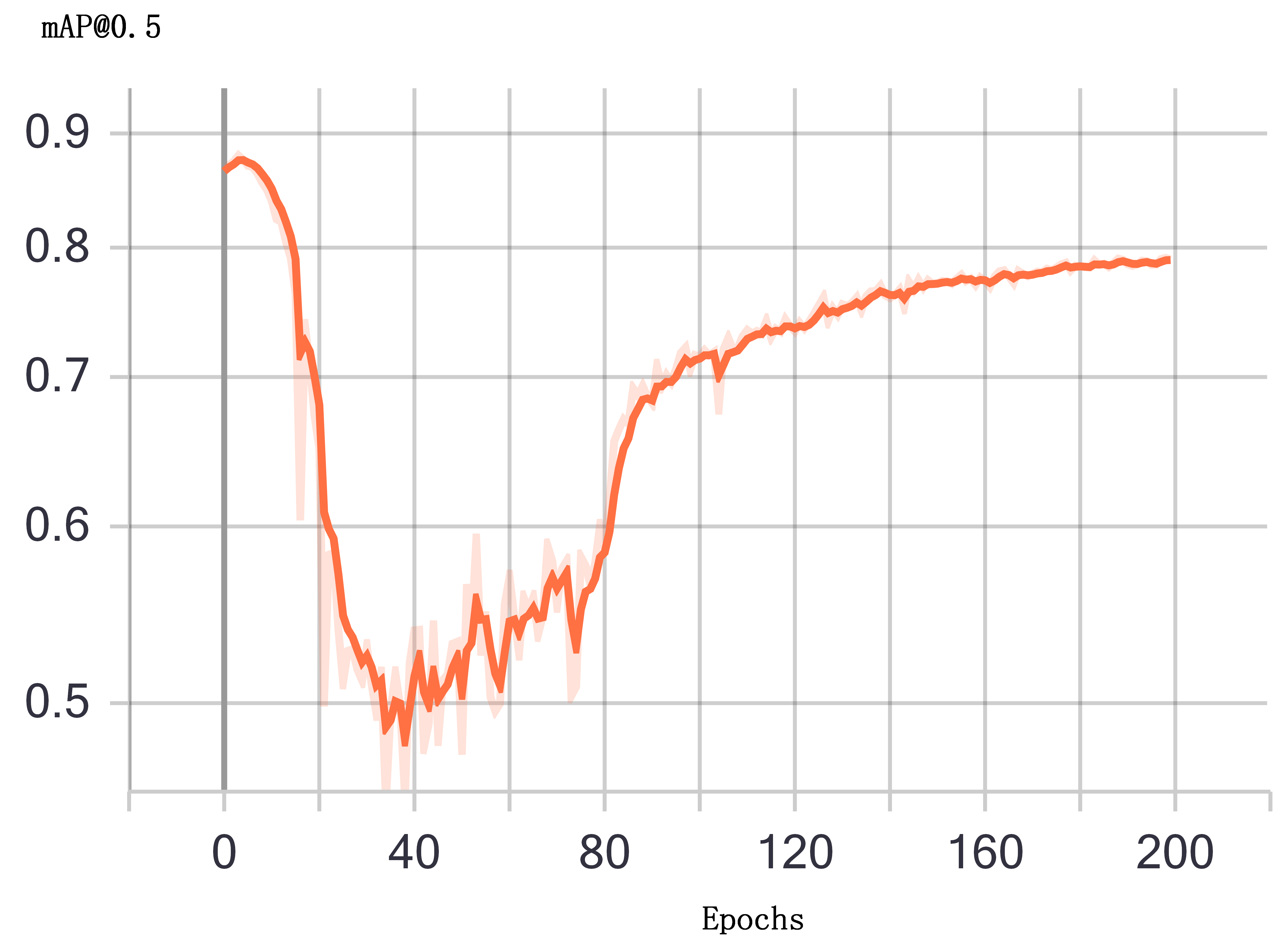}}\subfigure[image 4]{
\label{Fig.sub.d}
\includegraphics[width=6.2cm,height = 3.8cm]{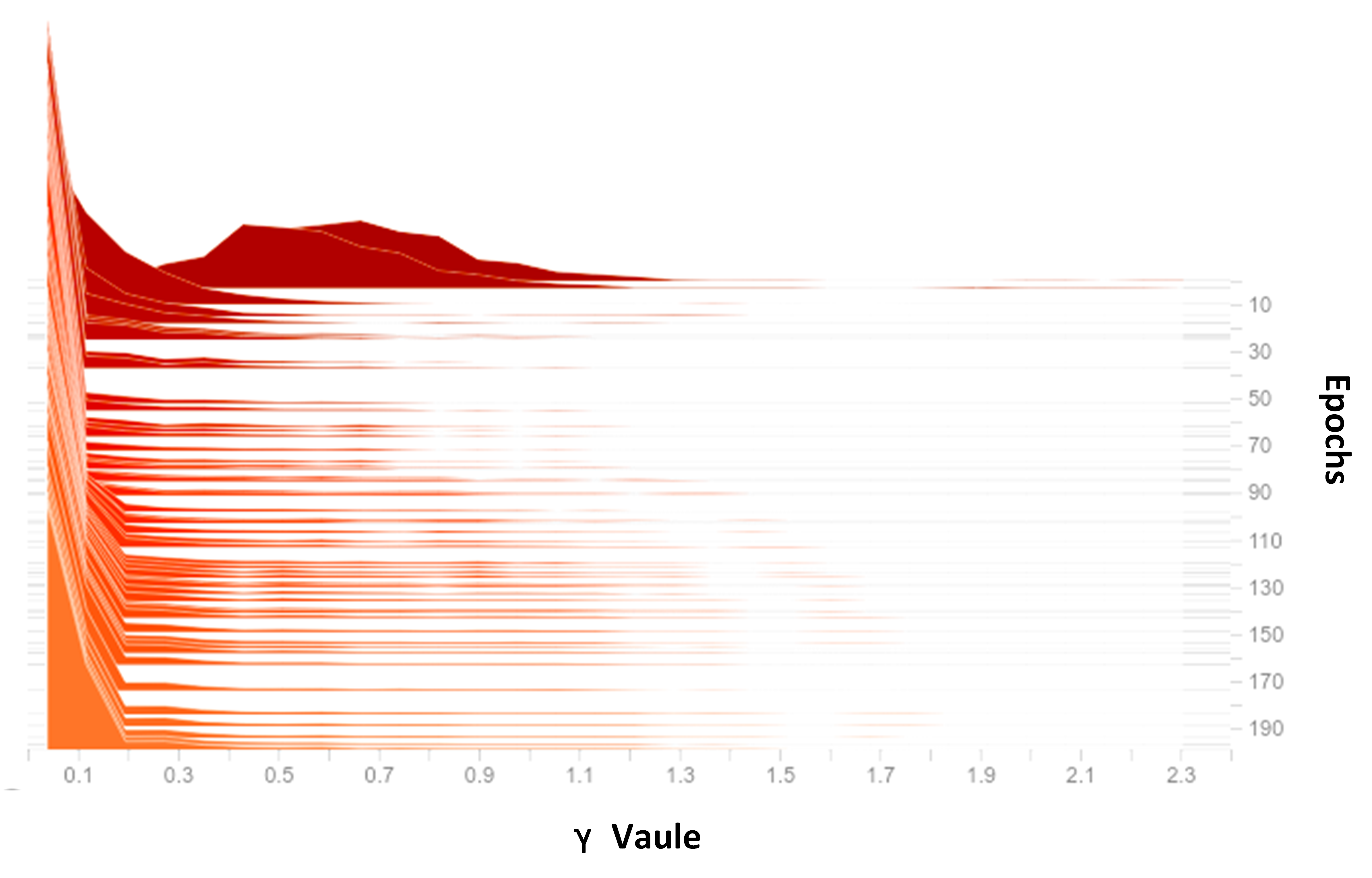}}

\caption{The top two figures show the process of the common sparse training(stable sparse rate $s$) and the bottom two figures show the process of the dynamic sparse training(variable sparse rate $s$). (a), (c) represents the accuracy of the  network changes with the training epochs and (b), (d)  represents the distribution of $\gamma$ coefficient changes with the training epochs.}
\label{Fig.11}
\end{figure}

\begin{table}[htbp]
\centering
\caption{Ablation Studies for sparse training and dynamic sparse training.}
\setlength{\tabcolsep}{0.5mm}{
\begin{tabular}{lccccc} 
\toprule
  Model  & Sparse training &  Dynamic Sparse training  &  mAP@0.5  & Model-size   \\
  \midrule
 CSPDarkNet-Yolov4  &  \checkmark &   &  71.2    & 256MB  \\
 CSPDarkNet-Yolov4  &  & \checkmark   &  79.8    & 256MB  \\
 DarkNet53-Yolov3   &  \checkmark &   &  59.2      & 246MB  \\
 DarkNet53-Yolov3   &  & \checkmark   &   66.5    & 246MB  \\
  MobileNet-Yolov3   & \checkmark &    &  71.3    &  95MB  \\
  MobileNet-Yolov3   &   & \checkmark  &  72.8  &  95MB   \\
\bottomrule
\end{tabular}
}
\label{Table 9}
\end{table}

For the  MobileNet-Yolov3,  DarkNet53-Yolov3, CSPDarknet-Yolov4, they have 96, 106, 162 network layers, respectively. It can be seen from the {Table 9}, the accuracy of the MobileNet-Yolov3,  DarkNet53-Yolov3, CSPDarknet-Yolov4 are 72.8, 66.5, 79.8, respectively. With the increase of the number detection network layers, the dynamic sparse training method has achieved a better trade-off between sparsity and accuracy compare to the common sparse training. Besides, when network layers are less than 100 layers, we notice that increasing the batch size also can improve the trade-off between sparsity and accuracy. 

\subsection{Ablation Studies for Group Channel Pruning}

In the group channel pruning experiments, we chose the  dynamic sparse training model as the network to be pruned and use the same pruning proportion for the common pruning and group channel pruning. For the CSPDarknet-Yolov4, the pruning proportion was 40 $\%$. For the DarkNet53-Yolov3 and MobileNet-Yolov3, the pruning proportion was 64 $\%$  and 65 $\%$, respectively.

\begin{table}[htbp]
\centering
\caption{Ablation Studies for common pruning\cite{liu2017learning} and group channel pruning}
\setlength{\tabcolsep}{0.5mm}{
\begin{tabular}{lccccc} 
\toprule
  Model  &  Pruning &  Group channel pruning  &  mAP@0.5  & Model-size   \\
  \midrule
  CSPDarkNet-Yolov4  & \checkmark &            &    48.5      & 94 MB      \\
  CSPDarkNet-Yolov4  &            & \checkmark &    56.2      & 90 MB      \\
  DarkNet53-Yolov3   & \checkmark &            &    60.9      & 62 MB     \\
  DarkNet53-Yolov3   &            & \checkmark &    65.2      & 21 MB        \\
  MobileNet-Yolov3   & \checkmark &            &    71.3      & 17 MB    \\
  MobileNet-Yolov3   &            & \checkmark &    72.2      & 15 MB       \\
\bottomrule
\end{tabular}
}
\label{Table 10}
\end{table}

It can be seen from the {Table 10},  comparing with the common pruning method \cite{liu2017learning}, our group channel pruning method can achieve a better balance between  model's size and accuracy for different detection networks. 

\subsection{Ablation Studies for Group Spatial Attention Distillation}

During the group spatial  experiments, we choose the original network as the teacher network, the pruned network (through the group channel pruning method) as the student network. 

For the CSPDarknet-Yolov4 and Mobilenet-Yolov3,  we extracted the spatial attention information  at specific scale feature maps from five groups as the knowledge, the scales of feature maps are 208 × 208, 104 × 104, 52 × 52, 26 × 26 and 13 × 13, respectively.
For the  Darknet53-Yolov3,  we  extracted the spatial attention information  at  104 × 104, 52 × 52, 26 × 26, and 13 × 13  scale feature maps from five groups.

\begin{table}[htbp]
\centering
\caption{Ablation Studies for fine tuning and group spatial attention Distillation (denoted as  GSA)}
\setlength{\tabcolsep}{0.5mm}{
\begin{tabular}{lccccc} 
\toprule
  Model  & Fine tune &  GSA Distilling  &  mAP@0.5  & Model-size   \\
  \midrule
  CSPDarkNet-Yolov4  & \checkmark &            &    81.1      & 90 MB      \\
  CSPDarkNet-Yolov4  &            & \checkmark &    86.6      & 90 MB      \\
  DarkNet53-Yolov3   & \checkmark &            &    66.8      & 22 MB        \\
  DarkNet53-Yolov3   &            & \checkmark &    68.2      & 22 MB        \\
  MobileNet-Yolov3   & \checkmark &            &         72.1 & 15 MB       \\
  MobileNet-Yolov3   &            & \checkmark &         73.2 & 15 MB       \\
\bottomrule
\end{tabular}
}
\label{Table }
\end{table}

As shown in {Table 11}, comparing to fine tune the pruned network, group spatial attention distillation can achieve a better  accuracy for different detection networks.

In addition, we qualitatively demonstrate the effectiveness of the pruned network through group spatial attention distilling. As shown in {Fig.13},  (a),(b),(c) show  the original  CSPDarknet-Yolov4, DarkNet53-Yolov3, MobileNet-Yolov3, detection results, respectively. And (d),(e),(f) show the detection results of the pruned network.

\begin{figure}[t]
\centering  
\subfigure[Image 1]{
\label{Fig.sub.1}
\includegraphics[width=5.2cm,height = 3.8cm]{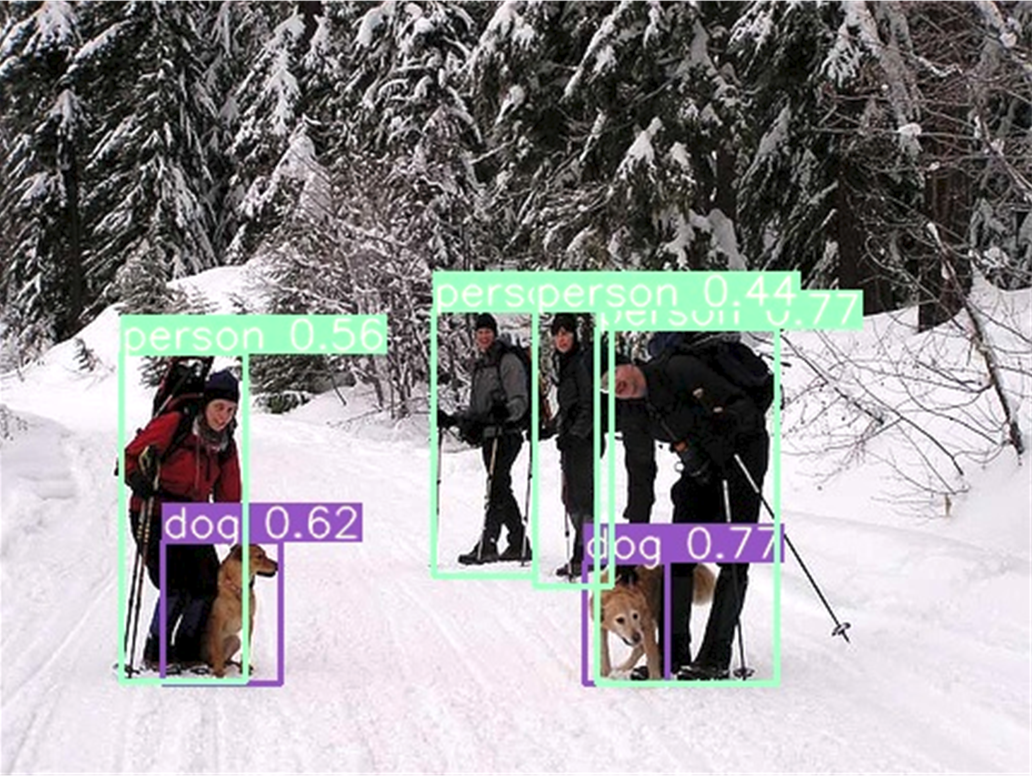}}\subfigure[Image 2]{
\label{Fig.sub.2}
\includegraphics[width=5.2cm,height = 3.8cm]{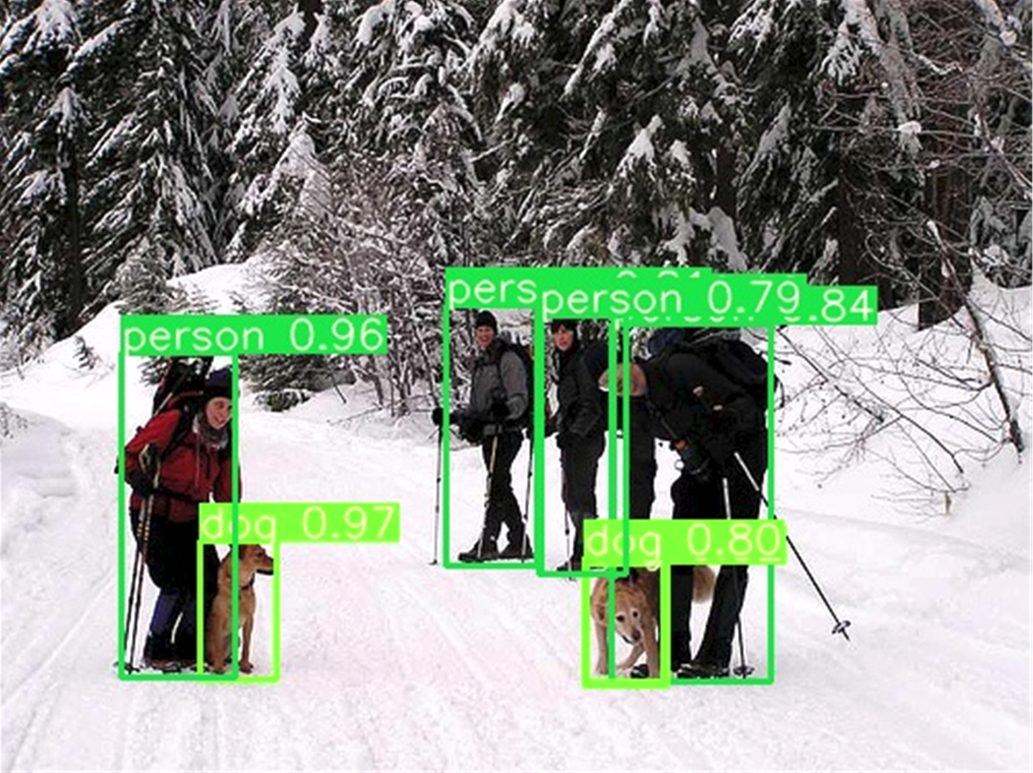}}\subfigure[Image 3]{
\label{Fig.sub.2}
\includegraphics[width=5.2cm,height = 3.8cm]{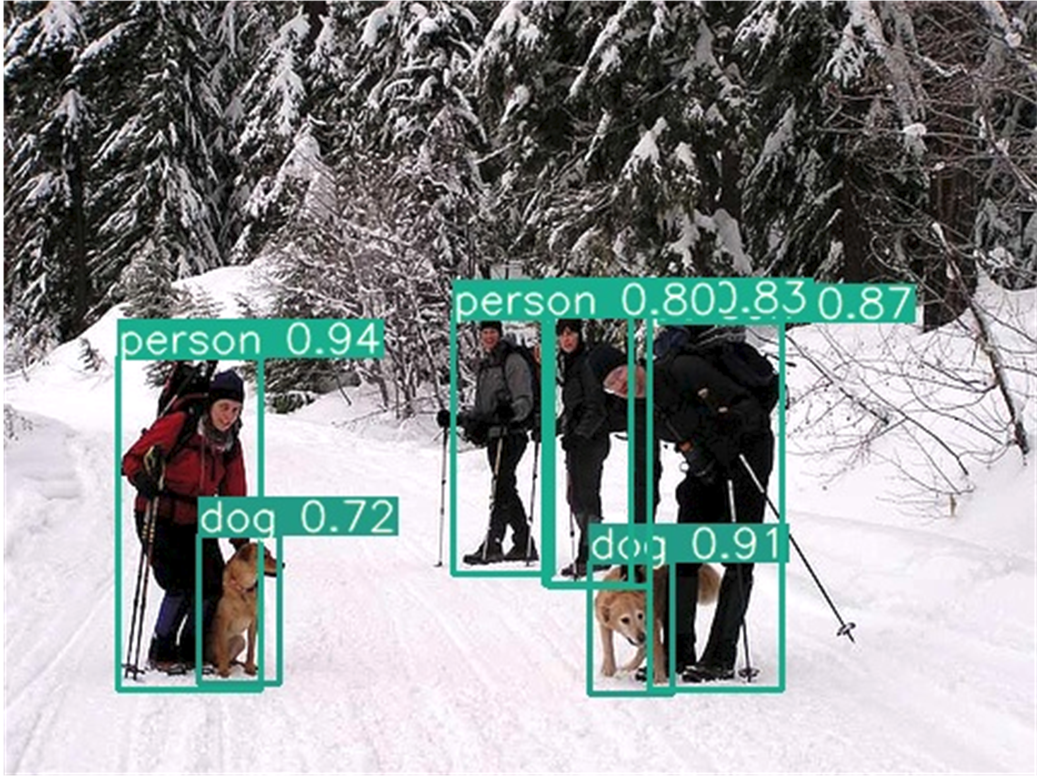}}

\subfigure[Image 4]{
\label{Fig.sub.4}
\includegraphics[width=5.2cm,height = 3.8cm]{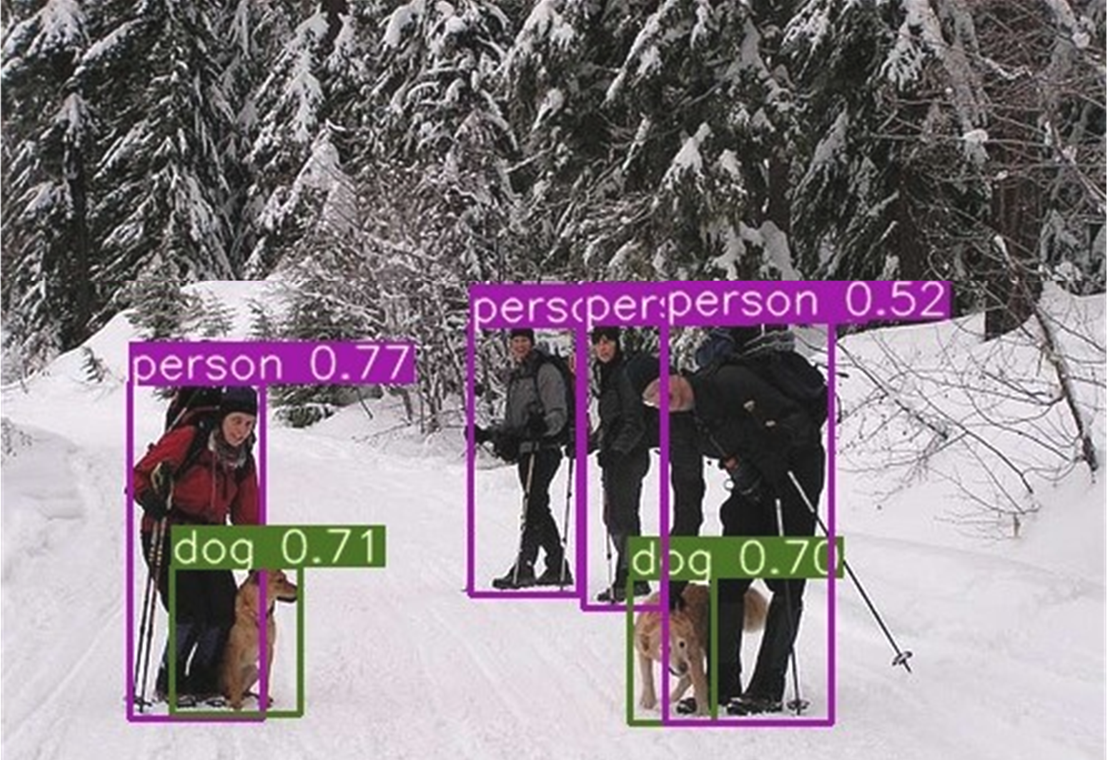}}\subfigure[Image 5]{
\label{Fig.sub.5}
\includegraphics[width=5.2cm,height = 3.8cm]{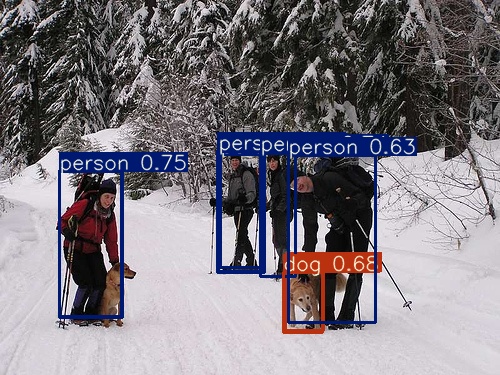}}\subfigure[Image 6]{
\label{Fig.sub.6}
\includegraphics[width=5.2cm,height = 3.8cm]{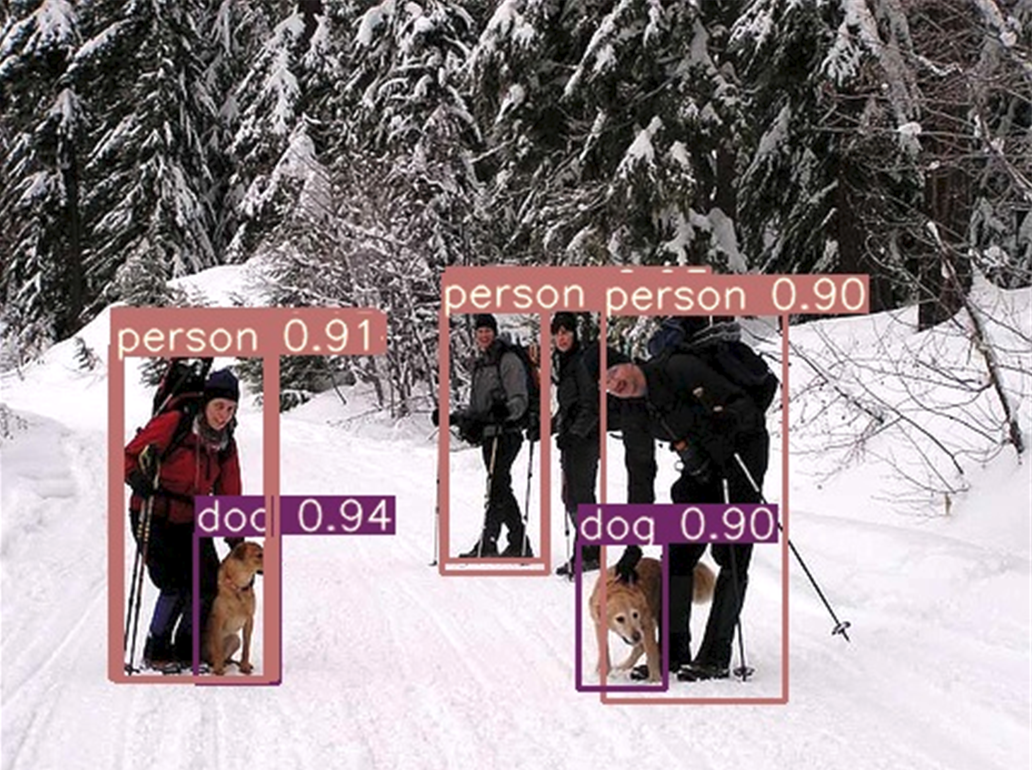}}

\caption{The top three figures show the detection results of the original CSPDarkNet-Yolov4(a), DarkNet-Yolov3(b), MobileNet-Yolov3(c) network, and the bottom three figures show the detection results of the pruned CSPDarkNet-Yolov4(d), DarkNet-Yolov3(e), MobileNet-Yolov3(f) network.}
\label{Fig.13}
\end{figure}

\section{Conclusions}

In this paper, we present a three-stage model compression approach for the object detection network, which is dynamic sparse training, group channel pruning, and spatial attention distillation. Firstly, we introduce dynamic sparse training to select out the insignificant channels in the layers and maintain a good balance of networks' sparsity and accuracy. Next, we propose a group channel pruning method. Under the same pruning rate, our group pruning method has less influence on the accuracy of the network and can obtain considerable model compression comparing with other pruning methods. After that, we extract each group's spatial attention information as the knowledge for distillation. Compared with the direct fine-tuning of the pruned model, our group spatial attention distilling method can recover the pruned network to higher accuracy. Furthermore, we deploy the compressed model on the edge device Jetson Nano to demonstrate that our method  can be directly deployed without the support of special hardware or software and can achieve the acceleration effect. To demonstrate the generality and effectiveness of our proposed approach, in our experiments, we replace the backbone to  MobileNet, DarkNet53, and CSPDarknet to construct the detection network and then use our proposed methods, the experimental results are satisfactory. We believe that the proposed methodology and approach are promising to be evaluated for compressing other object detection networks.

\begin{acknowledgements}
This work was supported in part by the National Natural Science Foundation of China under Grant 61961014, 61963012 and the Hainan Provincial Natural Science Foundation of China under Grant  620RC556, 620RC564.
\end{acknowledgements}

%
%

\bibliographystyle{ieeetr}       
\bibliography{cas-refs}   

\begin{thebibliography}{10}

\bibitem{sullivan2018deep}
D.~P. Sullivan, C.~F. Winsnes, L.~{\AA}kesson, M.~Hjelmare, M.~Wiking,
  R.~Schutten, L.~Campbell, H.~Leifsson, S.~Rhodes, A.~Nordgren, {\em et~al.},
  ``Deep learning is combined with massive-scale citizen science to improve
  large-scale image classification,'' {\em Nature biotechnology}, vol.~36,
  no.~9, pp.~820--828, 2018.

\bibitem{liu2020deep}
L.~Liu, W.~Ouyang, X.~Wang, P.~Fieguth, J.~Chen, X.~Liu, and
  M.~Pietik{\"a}inen, ``Deep learning for generic object detection: A survey,''
  {\em International journal of computer vision}, vol.~128, no.~2,
  pp.~261--318, 2020.

\bibitem{sultana2020evolution}
F.~Sultana, A.~Sufian, and P.~Dutta, ``Evolution of image segmentation using
  deep convolutional neural network: a survey,'' {\em Knowledge-Based Systems},
  vol.~201, p.~106062, 2020.

\bibitem{krizhevsky2017imagenet}
A.~Krizhevsky, I.~Sutskever, and G.~E. Hinton, ``Imagenet classification with
  deep convolutional neural networks,'' {\em Communications of the ACM},
  vol.~60, no.~6, pp.~84--90, 2017.

\bibitem{he2016deep}
K.~He, X.~Zhang, S.~Ren, and J.~Sun, ``Deep residual learning for image
  recognition,'' in {\em Proceedings of the IEEE conference on computer vision
  and pattern recognition}, pp.~770--778, 2016.

\bibitem{howard2018inverted}
A.~Howard, A.~Zhmoginov, L.-C. Chen, M.~Sandler, and M.~Zhu, ``Inverted
  residuals and linear bottlenecks: Mobile networks for classification,
  detection and segmentation,'' 2018.

\bibitem{ren2016faster}
S.~Ren, K.~He, R.~Girshick, and J.~Sun, ``Faster r-cnn: towards real-time
  object detection with region proposal networks,'' {\em IEEE transactions on
  pattern analysis and machine intelligence}, vol.~39, no.~6, pp.~1137--1149,
  2016.

\bibitem{liu2016ssd}
W.~Liu, D.~Anguelov, D.~Erhan, C.~Szegedy, S.~Reed, C.-Y. Fu, and A.~C. Berg,
  ``Ssd: Single shot multibox detector,'' in {\em European conference on
  computer vision}, pp.~21--37, Springer, 2016.

\bibitem{farhadi2018yolov3}
A.~Farhadi and J.~Redmon, ``Yolov3: An incremental improvement,'' in {\em
  Computer Vision and Pattern Recognition}, pp.~1804--02767, 2018.

\bibitem{bochkovskiy2020yolov4}
A.~Bochkovskiy, C.-Y. Wang, and H.-Y.~M. Liao, ``Yolov4: Optimal speed and
  accuracy of object detection,'' {\em arXiv preprint arXiv:2004.10934}, 2020.

\bibitem{han2016eie}
S.~Han, X.~Liu, H.~Mao, J.~Pu, A.~Pedram, M.~A. Horowitz, and W.~J. Dally,
  ``Eie: Efficient inference engine on compressed deep neural network,'' {\em
  ACM SIGARCH Computer Architecture News}, vol.~44, no.~3, pp.~243--254, 2016.

\bibitem{rathi2018stdp}
N.~Rathi, P.~Panda, and K.~Roy, ``Stdp-based pruning of connections and weight
  quantization in spiking neural networks for energy-efficient recognition,''
  {\em IEEE Transactions on Computer-Aided Design of Integrated Circuits and
  Systems}, vol.~38, no.~4, pp.~668--677, 2018.

\bibitem{abderrahmane2020design}
N.~Abderrahmane, E.~Lemaire, and B.~Miramond, ``Design space exploration of
  hardware spiking neurons for embedded artificial intelligence,'' {\em Neural
  Networks}, vol.~121, pp.~366--386, 2020.

\bibitem{luo2020autopruner}
J.-H. Luo and J.~Wu, ``Autopruner: An end-to-end trainable filter pruning
  method for efficient deep model inference,'' {\em Pattern Recognition},
  vol.~107, p.~107461, 2020.

\bibitem{fernandes2021pruning}
F.~E. Fernandes~Jr and G.~G. Yen, ``Pruning deep convolutional neural networks
  architectures with evolution strategy,'' {\em Information Sciences},
  vol.~552, pp.~29--47, 2021.

\bibitem{cheng2021intelligent}
Y.~Cheng, M.~Lin, J.~Wu, H.~Zhu, and X.~Shao, ``Intelligent fault diagnosis of
  rotating machinery based on continuous wavelet transform-local binary
  convolutional neural network,'' {\em Knowledge-Based Systems}, vol.~216,
  p.~106796, 2021.

\bibitem{deng2018gxnor}
L.~Deng, P.~Jiao, J.~Pei, Z.~Wu, and G.~Li, ``Gxnor-net: Training deep neural
  networks with ternary weights and activations without full-precision memory
  under a unified discretization framework,'' {\em Neural Networks}, vol.~100,
  pp.~49--58, 2018.

\bibitem{tung2018deep}
F.~Tung and G.~Mori, ``Deep neural network compression by in-parallel
  pruning-quantization,'' {\em IEEE transactions on pattern analysis and
  machine intelligence}, vol.~42, no.~3, pp.~568--579, 2018.

\bibitem{hu2021opq}
P.~Hu, X.~Peng, H.~Zhu, M.~M.~S. Aly, and J.~Lin, ``Opq: Compressing deep
  neural networks with one-shot pruning-quantization,'' in {\em Proceedings of
  the Thirty-Fifth AAAI Conference on Artificial Intelligence (AAAI-21),
  Vancouver, VN, Canada}, pp.~2--9, 2021.

\bibitem{hinton2015distilling}
G.~Hinton, O.~Vinyals, and J.~Dean, ``Distilling the knowledge in a neural
  network,'' {\em stat}, vol.~1050, p.~9, 2015.

\bibitem{xu2019lightweightnet}
T.-B. Xu, P.~Yang, X.-Y. Zhang, and C.-L. Liu, ``Lightweightnet: Toward fast
  and lightweight convolutional neural networks via architecture
  distillation,'' {\em Pattern Recognition}, vol.~88, pp.~272--284, 2019.

\bibitem{zhang2021adversarial}
H.~Zhang, Z.~Hu, W.~Qin, M.~Xu, and M.~Wang, ``Adversarial co-distillation
  learning for image recognition,'' {\em Pattern Recognition}, vol.~111,
  p.~107659, 2021.

\bibitem{song2021classifier}
D.~Song, J.~Xu, J.~Pang, and H.~Huang, ``Classifier-adaptation knowledge
  distillation framework for relation extraction and event detection with
  imbalanced data,'' {\em Information Sciences}, vol.~573, pp.~222--238, 2021.

\bibitem{wang2021joint}
Z.-R. Wang and J.~Du, ``Joint architecture and knowledge distillation in cnn
  for chinese text recognition,'' {\em Pattern Recognition}, vol.~111,
  p.~107722, 2021.

\bibitem{li2021mutual}
Z.~Li, Y.~Ming, L.~Yang, and J.-H. Xue, ``Mutual-learning sequence-level
  knowledge distillation for automatic speech recognition,'' {\em
  Neurocomputing}, vol.~428, pp.~259--267, 2021.

\bibitem{shen2020knowledge}
P.~Shen, X.~Lu, S.~Li, and H.~Kawai, ``Knowledge distillation-based
  representation learning for short-utterance spoken language identification,''
  {\em IEEE/ACM Transactions on Audio, Speech, and Language Processing},
  vol.~28, pp.~2674--2683, 2020.

\bibitem{yang2021partially}
M.~Yang, Y.~Li, Z.~Huang, Z.~Liu, P.~Hu, and X.~Peng, ``Partially view-aligned
  representation learning with noise-robust contrastive loss,'' in {\em
  Proceedings of the IEEE/CVF Conference on Computer Vision and Pattern
  Recognition}, pp.~1134--1143, 2021.

\bibitem{ibrahem2021real}
H.~Ibrahem, A.~D.~A. Salem, and H.-S. Kang, ``Real-time weakly supervised
  object detection using center-of-features localization,'' {\em IEEE Access},
  vol.~9, pp.~38742--38756, 2021.

\bibitem{zhou2021rsanet}
Q.~Zhou, J.~Wang, J.~Liu, S.~Li, W.~Ou, and X.~Jin, ``Rsanet: Towards real-time
  object detection with residual semantic-guided attention feature pyramid
  network,'' {\em Mobile Networks and Applications}, vol.~26, no.~1,
  pp.~77--87, 2021.

\bibitem{zhou2022contextual}
Q.~Zhou, X.~Wu, S.~Zhang, B.~Kang, Z.~Ge, and L.~J. Latecki, ``Contextual
  ensemble network for semantic segmentation,'' {\em Pattern Recognition},
  vol.~122, p.~108290, 2022.

\bibitem{zhou2020aglnet}
Q.~Zhou, Y.~Wang, Y.~Fan, X.~Wu, S.~Zhang, B.~Kang, and L.~J. Latecki,
  ``Aglnet: Towards real-time semantic segmentation of self-driving images via
  attention-guided lightweight network,'' {\em Applied Soft Computing},
  vol.~96, p.~106682, 2020.

\bibitem{kim2021edge}
S.-W. Kim, K.~Ko, H.~Ko, and V.~C. Leung, ``Edge-network-assisted real-time
  object detection framework for autonomous driving,'' {\em IEEE Network},
  vol.~35, no.~1, pp.~177--183, 2021.

\bibitem{sandler2019mobilenetv2}
M.~Sandler, A.~Howard, M.~Zhu, A.~Zhmoginov, and L.-C. Chen, ``Mobilenetv2:
  Inverted residuals and linear bottlenecks,'' 2019.

\bibitem{litiny}
Y.~Li, J.~Li, W.~Lin, and J.~Li, ``Tiny-dsod: Lightweight object detection for
  resource-restricted usages,''

\bibitem{qin2019thundernet}
Z.~Qin, Z.~Li, Z.~Zhang, Y.~Bao, G.~Yu, Y.~Peng, and J.~Sun, ``Thundernet:
  Towards real-time generic object detection on mobile devices,'' in {\em
  Proceedings of the IEEE/CVF International Conference on Computer Vision},
  pp.~6718--6727, 2019.

\bibitem{wang2018pelee}
R.~J. Wang, X.~Li, and C.~X. Ling, ``Pelee: A real-time object detection system
  on mobile devices,'' {\em Advances in Neural Information Processing Systems},
  vol.~31, pp.~1963--1972, 2018.

\bibitem{liupath}
S.~Liu, L.~Qi, H.~Qin, J.~Shi, and J.~Jia, ``Path aggregation network for
  instance segmentation,''

\bibitem{luo2017thinet}
J.-H. Luo, J.~Wu, and W.~Lin, ``Thinet: A filter level pruning method for deep
  neural network compression,'' in {\em Proceedings of the IEEE international
  conference on computer vision}, pp.~5058--5066, 2017.

\bibitem{liu2017learning}
Z.~Liu, J.~Li, Z.~Shen, G.~Huang, S.~Yan, and C.~Zhang, ``Learning efficient
  convolutional networks through network slimming,'' in {\em Proceedings of the
  IEEE international conference on computer vision}, pp.~2736--2744, 2017.

\bibitem{jordao2020discriminative}
A.~Jordao, M.~Lie, and W.~R. Schwartz, ``Discriminative layer pruning for
  convolutional neural networks,'' {\em IEEE Journal of Selected Topics in
  Signal Processing}, vol.~14, no.~4, pp.~828--837, 2020.

\bibitem{liu2018rethinking}
Z.~Liu, M.~Sun, T.~Zhou, G.~Huang, and T.~Darrell, ``Rethinking the value of
  network pruning,'' in {\em International Conference on Learning
  Representations}, 2018.

\bibitem{li2020eagleeye}
B.~Li, B.~Wu, J.~Su, and G.~Wang, ``Eagleeye: Fast sub-net evaluation for
  efficient neural network pruning,'' in {\em European Conference on Computer
  Vision}, pp.~639--654, Springer, 2020.

\bibitem{zhou2020two}
J.~Zhou, S.~Zeng, and B.~Zhang, ``Two-stage knowledge transfer framework for
  image classification,'' {\em Pattern Recognition}, vol.~107, p.~107529, 2020.

\bibitem{jung2020knowledge}
J.-W. Jung, H.-S. Heo, H.-J. Shim, and H.-J. Yu, ``Knowledge distillation in
  acoustic scene classification,'' {\em IEEE Access}, vol.~8,
  pp.~166870--166879, 2020.

\bibitem{chen2018training}
G.~Chen, X.~Zhang, X.~Tan, Y.~Cheng, F.~Dai, K.~Zhu, Y.~Gong, and Q.~Wang,
  ``Training small networks for scene classification of remote sensing images
  via knowledge distillation,'' {\em Remote Sensing}, vol.~10, no.~5, p.~719,
  2018.

\bibitem{chen2017learning}
G.~Chen, W.~Choi, X.~Yu, T.~Han, and M.~Chandraker, ``Learning efficient object
  detection models with knowledge distillation,'' {\em Advances in neural
  information processing systems}, vol.~30, 2017.

\bibitem{komodakis2017paying}
N.~Komodakis and S.~Zagoruyko, ``Paying more attention to attention: improving
  the performance of convolutional neural networks via attention transfer,'' in
  {\em ICLR}, 2017.

\bibitem{liu2020search}
Y.~Liu, X.~Jia, M.~Tan, R.~Vemulapalli, Y.~Zhu, B.~Green, and X.~Wang, ``Search
  to distill: Pearls are everywhere but not the eyes,'' in {\em Proceedings of
  the IEEE/CVF Conference on Computer Vision and Pattern Recognition},
  pp.~7539--7548, 2020.

\bibitem{santurkar2018does}
S.~Santurkar, D.~Tsipras, A.~Ilyas, and A.~Mdry, ``How does batch normalization
  help optimization,'' in {\em Proceedings of the 32nd international conference
  on neural information processing systems}, pp.~2488--2498, 2018.

\bibitem{dasgupta2021performance}
R.~Dasgupta, Y.~S. Chowdhury, and S.~Nanda, ``Performance comparison of
  benchmark activation function relu, swish and mish for facial mask detection
  using convolutional neural network,'' in {\em Intelligent Systems},
  pp.~355--367, Springer, 2021.

\bibitem{liu2019modified}
Y.~Liu, X.~Wang, L.~Wang, and D.~Liu, ``A modified leaky relu scheme (mlrs) for
  topology optimization with multiple materials,'' {\em Applied Mathematics and
  Computation}, vol.~352, pp.~188--204, 2019.

\bibitem{huang2020dc}
Z.~Huang, J.~Wang, X.~Fu, T.~Yu, Y.~Guo, and R.~Wang, ``Dc-spp-yolo: Dense
  connection and spatial pyramid pooling based yolo for object detection,''
  {\em Information Sciences}, vol.~522, pp.~241--258, 2020.

\bibitem{everingham2015pascal}
M.~Everingham, S.~A. Eslami, L.~Van~Gool, C.~K. Williams, J.~Winn, and
  A.~Zisserman, ``The pascal visual object classes challenge: A
  retrospective,'' {\em International journal of computer vision}, vol.~111,
  no.~1, pp.~98--136, 2015.

\bibitem{padilla2020survey}
R.~Padilla, S.~L. Netto, and E.~A. da~Silva, ``A survey on performance metrics
  for object-detection algorithms,'' in {\em 2020 International Conference on
  Systems, Signals and Image Processing (IWSSIP)}, pp.~237--242, IEEE, 2020.

\end{thebibliography}


\end{document}